\pdfoutput=1
\documentclass[11pt]{article}

\newif\ifarxiv
\arxivtrue % make it 'true' or 'false' if you want to keep the anonymized or public version

\ifarxiv
    \usepackage{acl}
\else
    \usepackage[review]{acl}
\fi
\usepackage{csquotes}
\usepackage{times}
\usepackage{latexsym}
\usepackage{amsmath}
\usepackage[T1]{fontenc}
\usepackage[utf8]{inputenc}

\usepackage{microtype}

\usepackage{inconsolata}
\usepackage{enumitem}
\usepackage{array}
\usepackage[capitalise]{cleveref}
\usepackage{listings}
\usepackage{graphicx}
\usepackage{longtable}
\usepackage{booktabs}
\usepackage{soul,xcolor}
\usepackage{longtable}
\sethlcolor{yellow}
\usepackage{subcaption}
\usepackage{multirow}
\usepackage{bm}
 \usepackage[most]{tcolorbox}

\newtcolorbox{myquotebox}{
  colback=white!0, % Transparent background
  colframe=black, % Black frame
  rounded corners,
  boxrule=0.5pt, % Frame thickness
  title=Prompt:,
  left=2mm, % Left margin within the box
  right=2mm, % Right margin within the box
  top=1mm, % Top margin within the box
  bottom=1mm % Bottom margin within the box
}

\definecolor{goldenrod}{rgb}{0,0,0.8}
\definecolor{deepred}{rgb}{0.6,0,0}
\definecolor{deepgreen}{rgb}{0,0.5,0}
\definecolor{pink}{RGB}{219, 48, 122}
\definecolor{forestgreen}{RGB}{34,139,34}
\definecolor{goldenrod}{RGB}{218,165,32}
\definecolor{sepia}{RGB}{112,66,20}

\crefname{figure}{Figure}{Figures}
\crefname{table}{Table}{Tables}
\crefname{appendix}{Appendix}{Appendices}
\crefname{section}{Section}{Sections}
\crefname{equation}{Eq.}{Eqs.}

\newcommand\myparagraph[1]{
\vskip 0.05in 
\noindent{\bf {#1}}}
\newcommand*\samethanks[1][\value{footnote}]{\footnotemark[#1]}

\title{Towards Faithful and Robust LLM Specialists for Evidence-Based Question-Answering}

\author{
    Tobias Schimanski\textsuperscript{\rm 1}\thanks{Equal Contributions.}, 
    Jingwei Ni\textsuperscript{\rm 1,2}\samethanks,
    Mathias Kraus\textsuperscript{\rm 3},
    Elliott Ash\textsuperscript{\rm 2},
     Markus Leippold\textsuperscript{\rm 1,4}  \\
    \textsuperscript{\rm 1}University of Zürich \hspace{5mm}
    \textsuperscript{\rm 2}ETH Zürich \hspace{5mm}
    \textsuperscript{\rm 3}University of Regensburg \\
    \textsuperscript{\rm 4}Swiss Finance Institute (SFI) \\ 
    \texttt{\{tobias.schimanski, markus.leippold\}@bf.uzh.ch, \{jingni, ashe\}@ethz.ch} \\\texttt{mathias.kraus@informatik.uni-regensburg.de}
    }

\begin{document}
\maketitle
\begin{abstract}
% situation/background
Advances towards more faithful and traceable answers of Large Language Models (LLMs) are crucial for various research and practical endeavors. One avenue in reaching this goal is basing the answers on reliable sources. 
% problem
However, this Evidence-Based QA has proven to work insufficiently with LLMs in terms of citing the correct sources (source quality) and truthfully representing the information within sources (answer attributability).
% solution
In this work, we systematically investigate how to robustly fine-tune LLMs for better source quality and answer attributability. Specifically, we introduce a data generation pipeline with automated data quality filters, which can synthesize diversified high-quality training and testing data at scale. We further introduce four test sets to benchmark the robustness of fine-tuned specialist models.
% evaluation
Extensive evaluation shows that fine-tuning on synthetic data improves performance on both in- and out-of-distribution. %Evidence-Based QA cases. 
Furthermore, we show that data quality, which can be drastically improved by proposed quality filters, matters more than quantity in improving Evidence-Based QA.\footnote{
All our codes, LLM generations, and human annotations are accessible through \url{https://github.com/EdisonNi-hku/Robust_Evidence_Based_QA}.
}
\end{abstract}

\section{Introduction}

Large Language Models (LLMs) \citep{brown2020language,ouyang2022training,gpt4techreport,touvron2023llama2,anil2023palm} have become the center of many cutting-edge applications due to their generalisability and information processing abilities. A typical application of LLMs is in \textit{Evidence-Based Question Answering (QA)}, where LLMs are expected to answer questions based on provided sources and cite the sources accurately \citep[e.g.,][]{ni-etal-2023-chatreport, vaghefi2023chatclimate, cuiChatLawOpenSourceLegal2023, liuEvaluatingVerifiabilityGenerative2023}. By providing these additional sources, multiple shortcomings of standalone LLMs, such as hallucination \citep{Ji_2023} and limited knowledge capacity \citep{huLargeLanguageModels2023}, can be addressed, thereby enhancing answer traceability \citep{gaoRetrievalAugmentedGenerationLarge2024}.
\begin{figure}[t]
    \centering
	\includegraphics[width=0.45\textwidth]{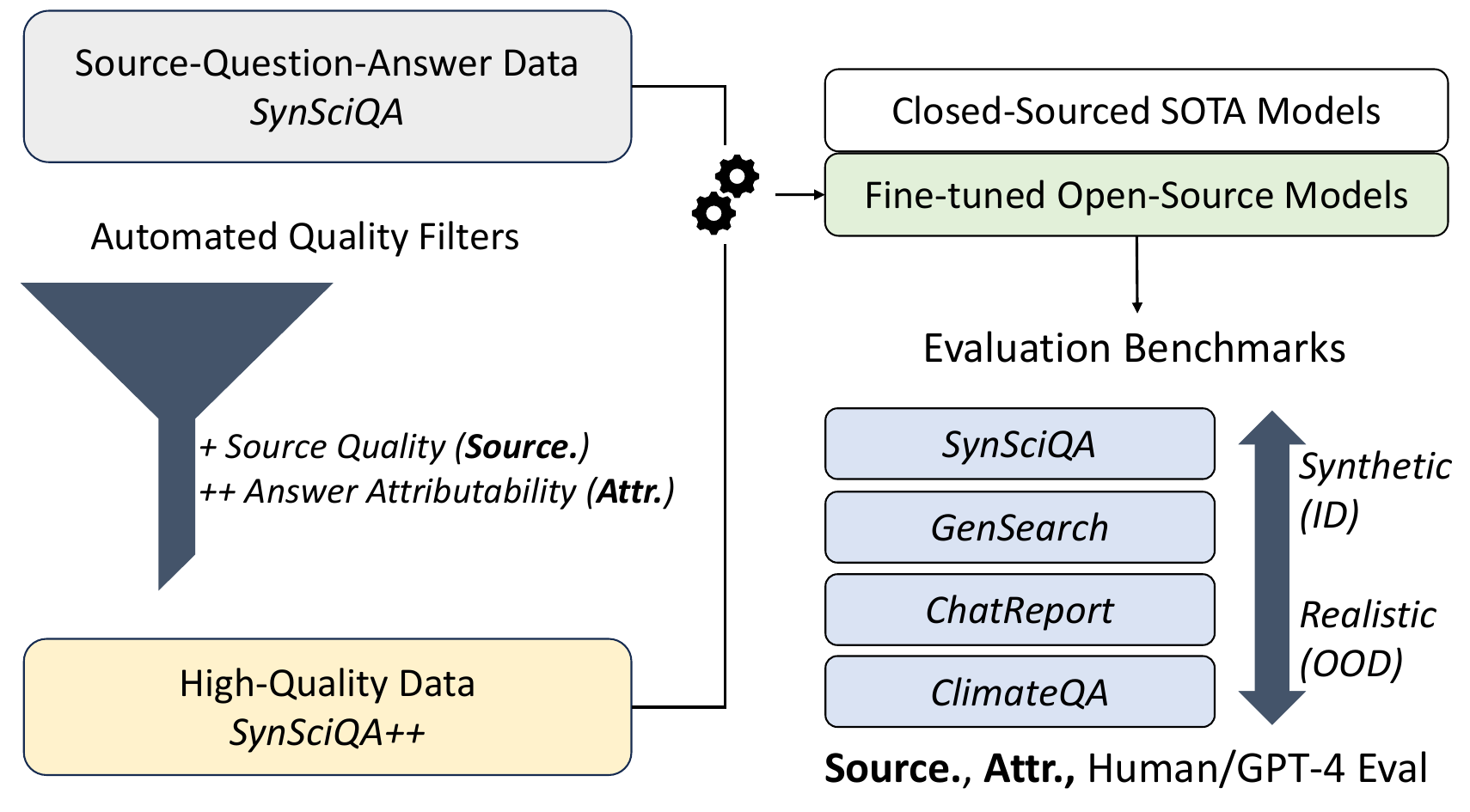}
	\caption{Synthetic data generation pipeline and Evaluation for Evidence-Based QA.}
	\label{fig:overview}
 % \vspace{-0.7em}
\end{figure}
However, the performance of existing LLMs on Evidence-Based QA is far from perfect. The SOTA close-sourced LLMs and generative search engines have an unignorable rate of hallucinated answers and false citation \citep{ni-etal-2023-chatreport,liuEvaluatingVerifiabilityGenerative2023}. Unfortunately, open-sourced LLMs are even less faithful than the already quality-lacking close-sourced LLMs in Evidence-Based QA (\citealp{yue-etal-2023-automatic}; \citealp{gao-etal-2023-enabling}; also see our evaluation in \cref{sec:zero-shot}), although they achieve competitive results on general instruction-following benchmarks \citep{touvron2023llama2,tunstallZephyrDirectDistillation2023}. We argue that this may prevent practitioners from building Evidence-Based QA (or other RAG) applications in a robust way. Therefore, efficient data creation and fine-tuning methods are urgently needed to improve LLMs' Evidence-Based QA performance in target applications.

To address this research gap, we first formulate quality dimensions for Evidence-Based QA. %as our optimization goal. 
Specifically, (1) LLMs need to \textbf{always} cite the right evidence at the end of each generated sentence to enable answer traceability, and (2) the answers need to be factually supported by the cited evidence. %This differs from previous work merely investigating the reliability of sentences if they had citations \citep{gao-etal-2023-enabling,yue-etal-2023-automatic}. We believe that providing complete citations is essential to make applications powered by LLMs more trustworthy and controllable.

Fine-tuning LLMs using Evidence-Based QA data that follow these quality dimensions seems straightforward. However, we identify two major challenges of fine-tuning LLMs into faithful evidence-based question answerers.
%\begin{enumerate}[itemsep=0pt,topsep=1pt,label=C\arabic*.]

\myparagraph{C1. Fine-Tuning Data Scalability}: Manual annotation for instruction tuning is costly \citep{DatabricksBlog2023DollyV2} and %which may prevent practitioners from annotating Evidence-Based QA data for their applications.
LLM-synthesized data can be a strong alternative \citep{yin-etal-2023-dynosaur}. However, the potentially lower quality of synthesized data may lead to suboptimal fine-tuning performance, given the SOTA LLMs' hallucination rate on Evidence-Based QA \citep{ni-etal-2023-chatreport,liuEvaluatingVerifiabilityGenerative2023}.

%\myparagraph{Auto-Evaluation for Faithfulness}: LLM generations are challenging to evaluated automatically for their faithfulness \citep{zhengJudgingLLMasajudgeMTBench2023}. Furthermore, there is no existing benchmark for Evidence-Based QA where faithfulness is pivotal. Thus, it is difficult for researchers and practitioners to validate and compare their models during application development.

\myparagraph{C2. Generalisability after Fine-tuning}: Previous work shows that diversified instruction tuning improves LLMs' generalisability \citep{chungScalingInstructionFinetunedLanguage2022,yin-etal-2023-dynosaur}. Hence, an intuitive worry is that fine-tuning LLMs (generalists) on Evidence-Based QA data (especially synthetic data) might turn LLMs into specialists that lack generalisability and, thus, struggle with out-of-distribution (OOD) questions and evidence.

% \myparagraph{Benchmarking Evidence-Based QA}: 
%\end{enumerate}

To address C1, we propose a data generation pipeline that synthesizes \textsc{SynSciQA} (\underline{Syn}thetic \underline{Sci}entific \underline{Q}uestion \underline{A}nswering), a well-diversified synthetic dataset for Evidence-Based QA, following prior work on data distillation for instruction tuning \citep[e.g.,][]{honovich-etal-2023-unnatural, tunstallZephyrDirectDistillation2023}. We further extend the pipeline with two novel quality filters to sift out low-quality synthetic data points, leading to \textsc{SynSciQA+} and \textsc{SynSciQA++} (see the left half of \cref{fig:overview}).
To address C2, we first collect an in-domain test set \textsc{SynSciQA}$_{test}$ with the data generation pipeline, which shares the data distribution with the training data (i.e., \textsc{SynSciQA}) but covers different topics. We further collect three test sets with different distances to the training data distribution to study the OOD performance (see the right half of \cref{fig:overview}). 

Extensive experiments on all proposed train and test settings show that (1) data quality is more important than quantity in Evidence-Based QA fine-tuning; (2) fine-tuning on generated data improves the performance on both in- and out-of-distribution test sets; and (3) performance scores on in-domain test set substantially indicate the OOD performance, suggesting that the synthetic data can be used for validation to estimate the OOD performance. All evaluation metrics are based on golden heuristics and best-performed models from previous work \citep{yue-etal-2023-automatic}, which we further verified with human and GPT-4 evaluation. In summary, our contributions include: 
\begin{enumerate}[itemsep=0pt,topsep=1pt]
\item We propose a data generation pipeline to obtain fine-tuning data for Evidence-Based QA in a salable way, which ensures data diversity and quality.
\item We propose four test sets to benchmark the in- and out-of-distribution performance of fine-tuned Evidence-Based QA specialists. 
\item We conduct an extensive evaluation to show that our data-synthesizing strategy leads to effective training and development set for Evidence-Based QA, and quality-filtering significantly improves fine-tuning performance.
\end{enumerate}

\section{Evidence-Based Question-Answering}
\begin{figure*}[t]
	\centering
	\includegraphics[width=0.95\textwidth]{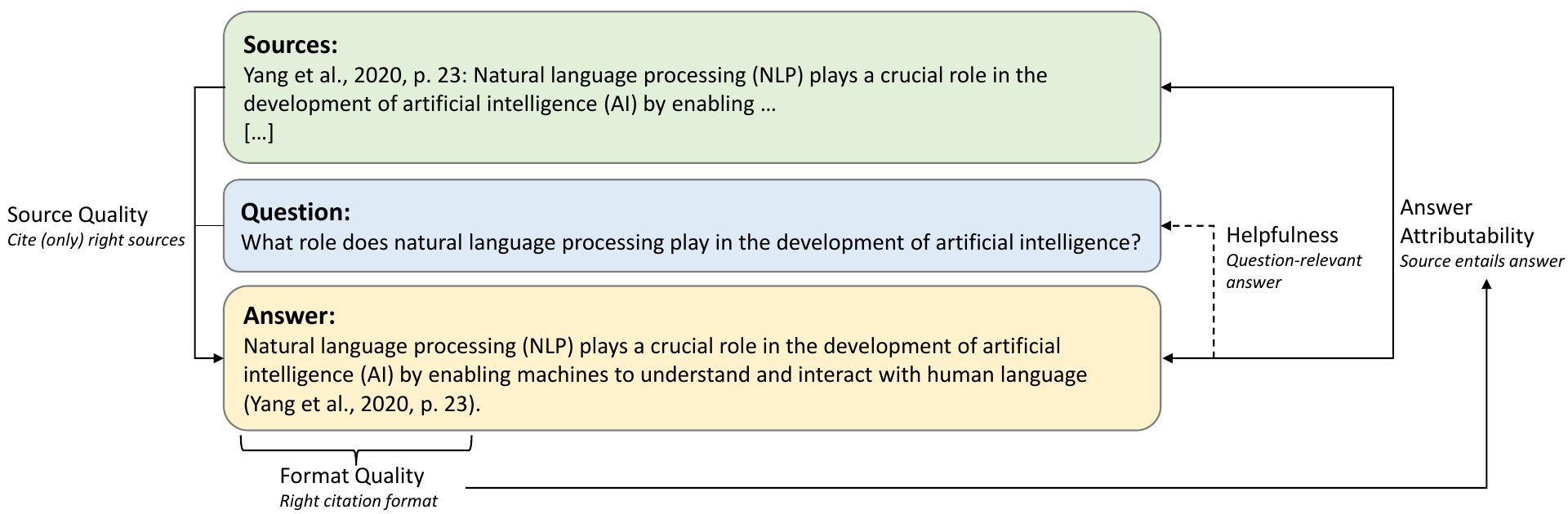}
	\caption{Quality Dimensions of Evidence-Based Question Answering.}
	\label{fig:qualitydims}
 % \vspace{-1em}
\end{figure*}

In this section, we formally define Evidence-Based QA. We further define its essential quality dimensions and the corresponding evaluation metrics.
% This project advances the traditional question-answering literature towards Evidence-Based QA. Therefore, in this section, we define this new task, quality dimensions, and methods to evaluate it.

\subsection{Task Definition}
The task in Evidence-Based QA represents answering a question based on provided sources while truthfully representing and citing the right sources. The model is presented with a set of zero or more \textit{relevant} $\mathcal{S}_{rel}$ and \textit{irrelevant} $\mathcal{S}_{irr}$ sources and a question $q$. Both are combined in a prompt template $\mathcal{P}$. The model $\mathcal{M}$ is expected to faithfully answer the question and support each answer sentence with a reference to given sources. That is, answer $\mathcal{A} = \mathcal{M}(\mathcal{P}( q, \mathcal{S}_{rel}
\cup \mathcal{S}_{irr})) = \{(a_{1}, s_{1}); (a_{2}, s_{2}); ...; (a_{n}, s_{n})\}$, where $n$ denotes the number of sentences in the answer $\mathcal{A}$; and $s_{i} \in \mathcal{S}_{cite}$ contains sources cited from $\mathcal{S}_{rel} \cup \mathcal{S}_{irr}$. All answer statements $a_{i}$ must be attributable to the cited sources $s_{i}$ rather than the model's parametric knowledge. The only scenario where the model is allowed to answer without citation is when the source evidence doesn't contain question-relevant information. However, the model should address this in its answer. %($\mathcal{A}_{\emptyset}$). %Thus, improving Evidence-Based QA centers around the optimization of the faithfulness and citation transparency of a model. 
Compared to the answer-attribution task defined in previous work \citep{liSurveyLargeLanguage2023}, Evidence-Based QA is more strict as it requires fully attributable and transparent answers.

\subsection{Quality Dimensions} \label{sec:22}
We focus on three pivotal quality dimensions to evaluate and improve Evidence-Based QA performance. (1) \textit{Source quality}. This describes whether the model's response only relies on \textit{relevant} sources, and, vice versa, does not include \textit{irrelevant} sources. (2) \textit{Format quality}, i.e., is a citation provided appropriately (to each sentence and in the right format) to maximize the traceability of the information? (3) \textit{Answer attributability}. Given correct citation format, an answer sentence is attributable only if it is entailed by the cited source and no hallucination or extrapolation is involved in answering the question. These quality dimensions are reflected in the following prompt template $\mathcal{P}$ which is constantly used in prompting and fine-tuning:
\begin{lstlisting}[frame=single, basicstyle=\ttfamily\scriptsize, xleftmargin=0pt, breaklines, numbers=none, xleftmargin=.05\columnwidth, xrightmargin=.05\columnwidth]
Given are the following sources: [BEGIN OF SOURCES]
{SOURCE_NAME_1}: {SOURCE_CONTENT_1}
{SOURCE_NAME_2}: {SOURCE_CONTENT_2}
...
{SOURCE_NAME_N}: {SOURCE_CONTENT_N} [END OF SOURCES] 

Can you respond to the question "{QUESTION}" by only relying on the sources. Ignore all sources that do not provide an answer to the question. 
Do not include any knowledge from outside of these sources. Only write a single paragraph. Each sentence must end with the reference in the form of (author, year, page number). Strictly follow this format. Citing multiple sources in one sentence is not allowed. 
However, if no source addresses the question, admit truthfully that no answer can be given.
Answer the question concisely and avoid being verbose.    
\end{lstlisting}

By ``\textsc{source name x}'' and ``\textsc{source content x}'', we denote the \textsc{x}-th source name and content correspondingly. ``\textsc{question}'' denotes the question to answer. Note that this prompt is not optimized with prompt engineering tricks. Hence, we hypothesize that our findings can be transferable to practitioners' use cases with different prompt templates.

Besides three quality dimensions, the prompt also requires that more than one citation for one statement is not allowed. We choose this design to maximize the answer traceability and enable clear judgments about attributability by both human and machine evaluators. The NLI models we use are trained on a one-claim-one-evidence setting \citep{yue-etal-2023-automatic} and thus may have suboptimal performance on multi-evidence claim verification, which is more challenging \citep{jiang-etal-2020-hover}.

Our quality dimensions and fine-tuning focus on faithfulness, the most significant shortcoming of open-sourced LLMs in Evidence-Based QA \citep{yue-etal-2023-automatic,gao-etal-2023-enabling}. Another important dimension is helpfulness, which can be defined as ``how well does the answer address the question?''. Our quality dimensions partially address helpfulness by measuring truthful responses based on question-relevant sources. However, we argue that helpfulness is hard to define and evaluate objectively. For this task, it is also challenging to disentangle helpfulness from faithfulness, as a response can only have high helpfulness if it follows the prompt well, i.e., obeying all the faithful citation requirements. To shed light on this aspect, we put additional analyses in Appendix \ref{appendix:helpfulnessEval}.

\subsection{Evaluation Metrics} \label{sec:metrics}
We propose two automated metrics using heuristics and automated models to evaluate these quality dimensions in Evidence-Based QA.

\myparagraph{Source quality score}: Given a prompt $\mathcal{P}(q, \mathcal{S}_{rel}
\cup \mathcal{S}_{irr})$ containing zero or more \textit{relevant} sources $\mathcal{S}_{rel}$ and \textit{irrelevant} sources $\mathcal{S}_{irr}$, the model outputs an answer $\mathcal{A}$ citing zero or more sources $\mathcal{S}_{cite}$. Then, the source quality of a sentence is a binary variable described by the following formula:
\begin{equation}
\resizebox{\columnwidth}{!}{%
$ % Start math mode
\begin{split}
SQ^{A} = 
\begin{cases} 
1, & \text{if } (|\mathcal{S}_{cite}| > 0) \land (\forall s_{i} \in \mathcal{S}_{cite}: s_{i} \not\in \mathcal{S}_{irr}) \\ 
1, & \text{if } (|\mathcal{S}_{cite}| = 0) \land (|\mathcal{S}_{rel}| = 0) \\
0, & \text{otherwise}.
\end{cases}
\end{split}
$ % End math mode
}
\end{equation}

In simple words, source quality equals one if no \textit{irrelevant} source is cited and if a non-zero amount \textit{relevant} sources is given, then the answer must contain a non-zero amount of citations. Otherwise, source quality equals zero.

\myparagraph{Attributability score}: 
Given an answer $\mathcal{A}$ with at least one citation, the attributability score of this instruction-answer pair can be calculated as:
\begin{equation}
    Attr.^{\mathcal{A}} = 1 - \frac{|\mathcal{A}_{un}| + |\mathcal{A}_{format}|}{|\mathcal{A}|} = \frac{|\mathcal{A}_{en}|}{|\mathcal{A}|}
\end{equation}
where $\mathcal{A}_{en}$ ($\mathcal{A}_{un}$) denotes the collection of factually entailed (unentailed) sentences, and $\mathcal{A}_{format}$ denotes the collection of answer sentences with a wrong format or without citation. %Then, we average the attributability scores of all instruction-answer pairs to obtain the overall attributability score for a system.
While the format quality is easy to measure through heuristics, the answer's sentence-source entailment is challenging and requires neural model prediction. In this work, we aggregate the best-performing attribution-prediction models of previous work: attrscore-flan-t5-xl and -xxl checkpoints from \citet{yue-etal-2023-automatic} to measure entailment. To achieve higher precision, a sentence is entailed by the cited source only if both models predict ``attributable''. The attributability score is not applicable for answers without any citation since models should not cite when there is no relevant source. Those answers are addressed by source quality scores (i.e., the model should cite when there is a relevant citation). We mostly follow ``citation recall'', a metric introduced by \citet{gao-etal-2023-enabling}, to design attributability scores but adjust it to our stricter setting of Evidence-Based QA (more details in \cref{appendix:faithfulnessDef}).

\section{Training Data Generation} \label{sec:data_gene_pipe}
Manually annotating Evidence-Based QA data that fulfills all quality dimensions is costly and lacks scalability. In this section, we introduce a novel data generation pipeline to obtain high-quality synthetic data. First, we use OpenAI LLMs to create a diverse and broad base data set of task-specific instruction-answer pairs (\textsc{SynSciQA}) following the structural approach of prior work for data distillation \citep[e.g.,][]{honovich-etal-2023-unnatural, tunstallZephyrDirectDistillation2023}. Second, we use our quality dimensions to create data sets of higher quality (\textsc{SynSciQA+} and \textsc{SynSciQA++}), enabling explorations on the importance of data quality \citep{zhou2023lima}.

\subsection{SynSciQA} \label{sect31}
We create \textsc{SynSciQA} leveraging both GPT-3.5 and GPT-4 to improve data diversity (GPT-3.5 contributes 75\%, see \cref{appendix:dataCreation} for more details). The data creation process proceeds in the following steps:
\begin{enumerate}[noitemsep]
\item Generate a broad array of 100+ scientific topics. 
\item Generate 25 distinctive questions for each topic.
\item Create three source paragraphs \textit{relevant} to each question.
\item Design an instruction encompassing 0-3 \textit{relevant} sources and 3-6 \textit{irrelevant} sources, along with the corresponding question (refer to the prompt template in \cref{sec:22}).
\item Create an answer to the question following the provided instruction.
\end{enumerate}

After the creation process, we split the data into a training set and a test set by topic. This allows us to test different topics in contrast to what we trained the model on and mitigates concerns about data leakage. Using this procedure leaves us with \textsc{SynSciQA} comprising 2143 training samples.\footnote{Some of the API requests were rejected, thus some topics have less than 25 questions.}

\subsection{Automated Quality Filters}
The clear task definition and data creation process allow us to apply quality filters on the dataset. First, we apply a \textit{source quality} filter to the original \textsc{SynSciQA}. Through its construction process, we know which sources are \textit{relevant} and \textit{irrelevant} to the question. Thus, we filter out data points that do not achieve a \textit{source quality} score of one. This leaves us with 1386 samples which we call \textsc{SynSciQA+}.

Second, we also apply the \textit{answer attributability} quality dimension as a filter to the dataset. A data point passes the attributability quality filter only if it obtains a full attributability score. This aims to ensure that answers are in the right format and entailed by the sources. Finally, our highest-quality \textsc{SynSciQA++} dataset contains 669 samples.

Since the entailment models in \textit{answer attributability} cannot be controlled with heuristics, we further perform a hand-annotation on 300 randomly-sampled source-answer pairs in the \textsc{SynSciQA++} dataset. Specifically, two annotators per sample investigate whether an answer sentence truthfully reflects the information in the referenced source. We find that both annotators agree on entailment for 94\% of the cases, are indecisive for 4\%, and conclusive about actual non-entailment for only 2\% of the cases (for more details, see Appendix \ref{appendix:handEvalTraining}). These results solidify the validity of the approach.

\section{Evaluation Datasets} \label{sec:eval_sets}
\begin{figure}[t]
	%\centering
	\includegraphics[width=0.4\textwidth]{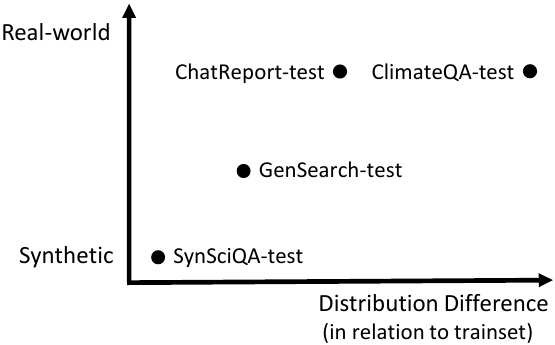}
	\caption{Evaluation Dataset's orientation towards real-world use case scenarios vs. their distribution's proximity to the trainsets.}
	\label{fig:data_distance}
 % \vspace{-0.7em}
\end{figure}

To assess the validity of the resulting models and their in- and out-of-distribution generalisability, we create a series of four evaluation datasets. The key differences are whether the data stems from a synthetic or real-world use case and how close the underlying data distribution is to the \textsc{SynSciQA} trainset. Our first evaluation benchmark is \textsc{SynSciQA}$_{test}$ which comprises 539 samples.

Our second evaluation benchmark is \textsc{GenSearch}$_{test}$. This is a dataset adapted from \citet{liu2023evaluating}. In this project, the authors create a dataset from posed questions to generative search engines and mark the question-relevant text part in the given source. We take this dataset and hand-evaluate 600 question-source pairs to distill full-text questions and clear corresponding sources that contain distinct variations of the information. This results in 276 question-source pairs or 106 questions with an average of 2.6 \textit{relevant} sources. After retaining this dataset, we can follow the creation process of \textsc{SynSciQA} (see Step 4 in section 3.1). See Appendix \ref{appendix:EvalBenchGenSearch} for more details.

We further create \textsc{ChatReport}$_{test}$. \textsc{ChatReport} is an open-source RAG tool that analyses companies' (sustainability) reports \citep{ni-etal-2023-chatreport}. It uses eleven sustainability-related questions to analyze the company's disclosure. Inherently, RAG systems' answers rely on source paragraphs from the underlying document, i.e. the company's report. Thus, we use the top-10 most relevant paragraphs (retrieved by \textsc{ChatReport} source code) as input for our system and create 110 instructions. This means we leave the structure of \textit{relevant} / \textit{irrelevant} sources and adopt a genuine RAG setting.

Finally, we use another RAG tool to create \textsc{ClimateQA}$_{test}$. ClimateQA\footnote{{https://huggingface.co/spaces/Ekimetrics/climate-question-answering}} is a RAG system that answers questions based on IPCC and IPBES reports. We pose 261 climate-related questions from \citet{welch2022what} to the system and store the outputted sources. Again, we use this data as input for our instruction form.

\cref{fig:data_distance} illustrates the distance between proposed test sets and \textsc{SynSciQA}$_{test}$ in dimensions of use case and data distribution. \textsc{ChatReport}$_{test}$ and \textsc{ClimateQA}$_{test}$ are directly extracted from real applications while \textsc{GenSearch}$_{test}$ is also from real research engine retrieval but with manual parsing (semi-synthetic). They also have different distribution distances to \textsc{SynSciQA}. \textsc{GenSearch}$_{test}$ contains vastly diversified, non-scientific questions; the source texts of \textsc{ChatReport}$_{test}$ contain formatting noise from sustainability reports; and \textsc{ClimateQA}$_{test}$ contains nested citations in its source texts, which may influence the models' citation correctness. Further explorations for each test set are showcased in \cref{appendix:test_example}.

\section{Experiments}
This section introduces our experiments and analyses in detail. We conduct experiments on Llama-2-chat-13b \citep{touvron2023llama2} and Zephyr-7b-$\beta$ \citep{tunstallZephyrDirectDistillation2023}. These models are chosen because they are from two widely used model families: Llama-2 and Mistral \citep{jiangMistral7B2023}. Their architecture can be representative of similar causal LLMs. %We do not choose larger models as 7B and 13B models are friendly to the majority of practitioners. 
We use aligned models instead of their base models (Llama-2-13b and Mistral-7b) to have models better understand the required quality dimensions for Evidence-Based QA. % A pilot comparison between aligned and unaligned models is shown in \cref{appendix:alignedornot}. 
We use QLoRA \citep{dettmers2023qlora} and greedy decoding for all LLM fine-tuning and inference correspondingly. Hyperparameters and other settings are presented in \cref{appendix:hyperparameter}.

\subsection{Zero-Shot Performance} \label{sec:zero-shot}
\begin{table*}[ht]
\small
\centering
\resizebox{\textwidth}{!}{
\begin{tabular}{lccccccccccc}
\toprule
& \multicolumn{2}{c}{\textbf{\textsc{SynSciQA}$_{test}$}} & & \multicolumn{2}{c}{\textbf{\textsc{GenSearch}$_{test}$}} & & \multicolumn{2}{c} {\textbf{\textsc{ChatReport}$_{test}$}} & & \multicolumn{2}{c} {\textbf{\textsc{ClimateQA}$_{test}$}} \\
\cline{2-3} \cline{5-6} \cline{8-9} \cline{11-12}  
& \textbf{Source.} & \textbf{Attr.} & & \textbf{Source.} & \textbf{Attr.}  & & \textbf{Source.} & \textbf{Attr.} & & \textbf{Source.} & \textbf{Attr.} \\
\toprule
Llama-2-13b-chat & $49.91$ & $25.01$ & & $69.80$ & $9.67$ & & - & $10.54$ & & - & $2.13$ \\
Zephyr-7b-$\beta$ & $36.92$ & $13.01$ & & $66.98$ & $5.29$  & & - & $5.22$ & & - & $2.30$\\
GPT-3.5 & $53.25$ & $64.93$ & & $96.23$ & $54.68$  & & - & $46.73$ & & - & $18.93$ \\
GPT-4 & $\mathbf{62.71}$ & $\mathbf{86.28}$ & & $\mathbf{99.06}$ & $\mathbf{60.34}$ & & - & $\mathbf{61.22}$ & & - & $\mathbf{28.01}$ \\

\bottomrule
\end{tabular}
}
\caption{Zero-shot performance of popular open- and close-sourced LLMs on proposed Evidence-Based QA benchmarks. Source. and Attr. are short for source quality, and attributability correspondingly. Source quality is not applicable for \textsc{ChatReport}$_{test}$ and \textsc{ClimateQA}$_{test}$ as source-relevance labels are not available for these real RAG systems (see discussions in \cref{appendix:relevance_label}).}
\label{tab:zero-shot}
% \vspace{-0.7em}
\end{table*}

We first use the proposed test sets and evaluation metrics to benchmark the zero-shot performance of close-sourced and open-sourced LLMs on Evidence-Based QA. The results are shown in \cref{tab:zero-shot}. We find that there is a significant performance gap between open- and close-sourced LLMs on Evidence-Based QA, although they achieve comparable performance on general instruction-following benchmarks (for instance, Zephyr-7b-$\beta$ vs. GPT-3.5 on MT-Bench \citep{tunstallZephyrDirectDistillation2023}). Similarly, \citet{tunstallZephyrDirectDistillation2023} shows that Zephyr-7b-$\beta$ outperforms Llama-2-70b-chat on all dimensions of MT-Bench, while our evaluation shows that Llama-2-13b-chat hallucinated less than Zephyr-7b-$\beta$ on Evidence-Based QA. Therefore, the proposed Evidence-Based QA benchmarks can be an effective resource to benchmark LLMs' faithfulness, supplementing MT-Bench.

All models achieve lower attributability scores on non-synthetic test sets, indicating that these more realistic settings are more challenging and current LLMs are far from faithful in Evidence-Based QA. The source quality scores on \textsc{GenSearch}$_{test}$ are relatively high since its questions and corresponding sources are extremely diversified (see \cref{appendix:EvalBenchGenSearch}). Thus, it is easier to tell whether a source is relevant to a question or not.

\subsection{\textsc{SynSciQA} Fine-Tuning} \label{sec:fine-tuning}

\begin{figure}[t]
	%\centering
	\includegraphics[width=\columnwidth]{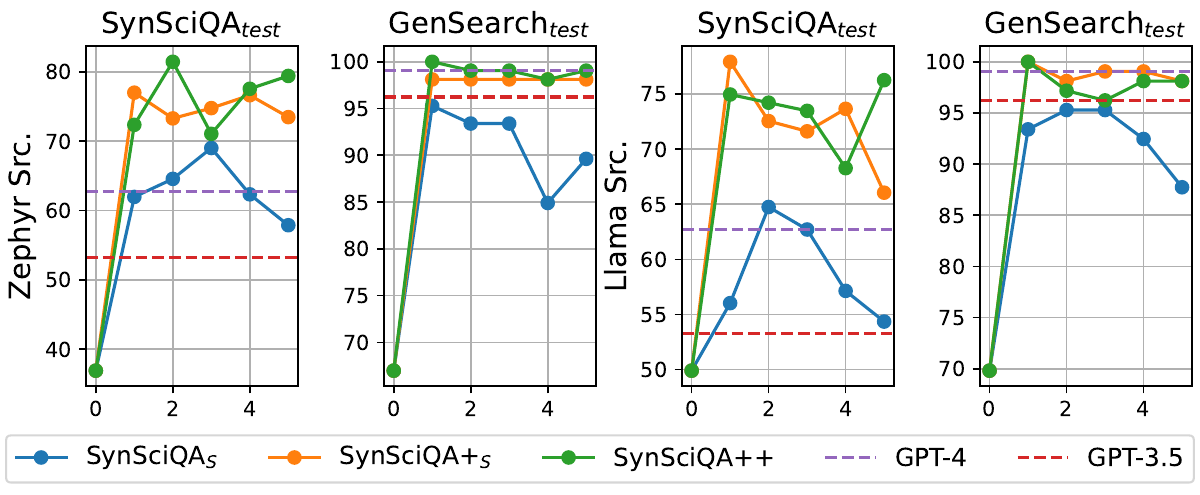}
        \captionsetup{skip=5pt}
	\caption{Controlling \textbf{quantity}, Source Quality scores vs. number of epoch, caused by different \textbf{quality}.}
	\label{fig:quality_source_all}
 % \vspace{-0.7em}
\end{figure}
\begin{figure}[t]
	%\centering
	\includegraphics[width=\columnwidth]{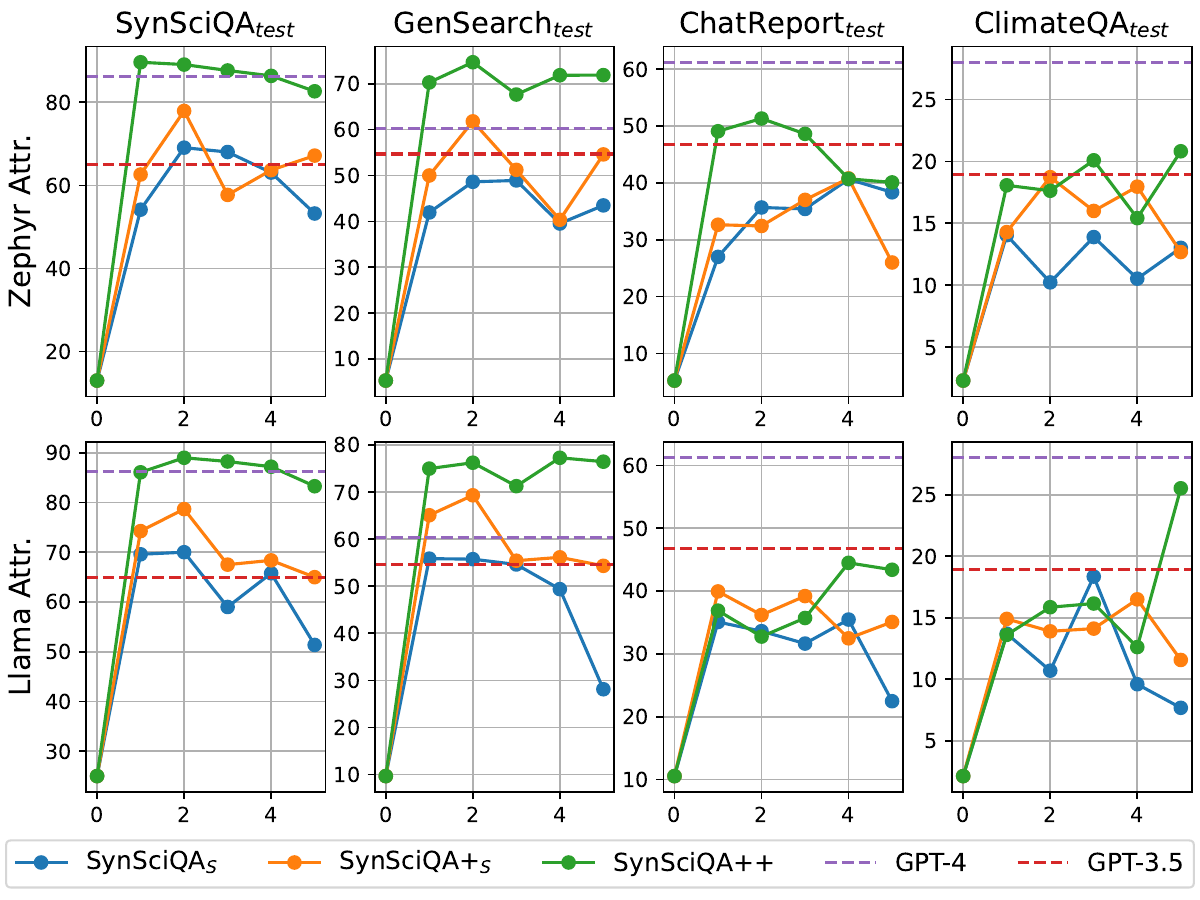}
         \captionsetup{skip=5pt}
	\caption{Controlling \textbf{quantity}, Attributability scores vs. number of epoch, caused by different \textbf{quality}.}
	\label{fig:quality_attr_all}
 % \vspace{-0.7em}
\end{figure}

\begin{figure}[t]
	%\centering
	\includegraphics[width=\columnwidth]{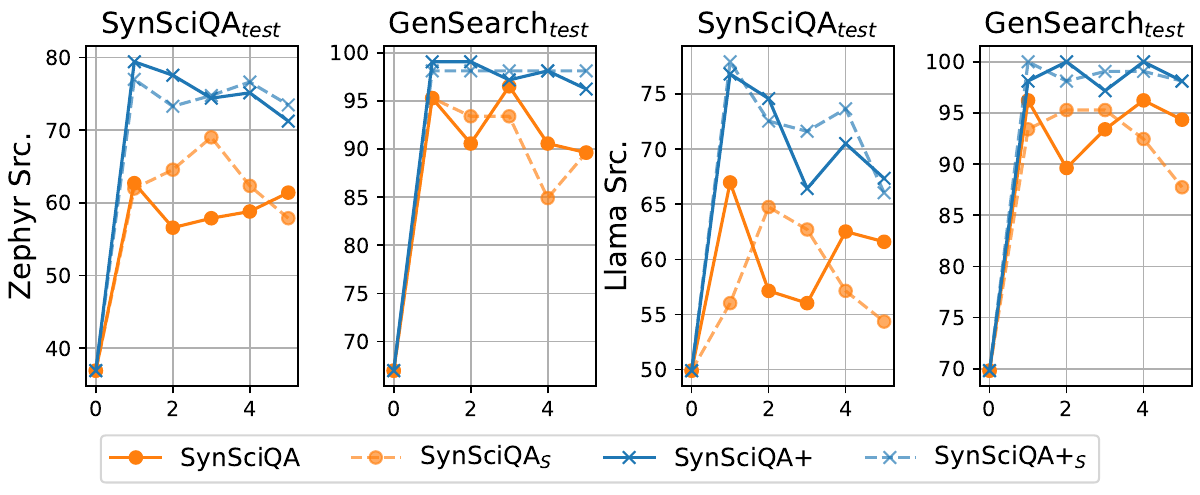}
        \captionsetup{skip=5pt}
	\caption{Controlling \textbf{quality}, Source Quality scores vs. number of epoch, caused by different \textbf{quantity}.}
	\label{fig:quantity_source_all}
 % \vspace{-0.8em}
\end{figure}

\begin{figure}[t]
	%\centering
	\includegraphics[width=\columnwidth]{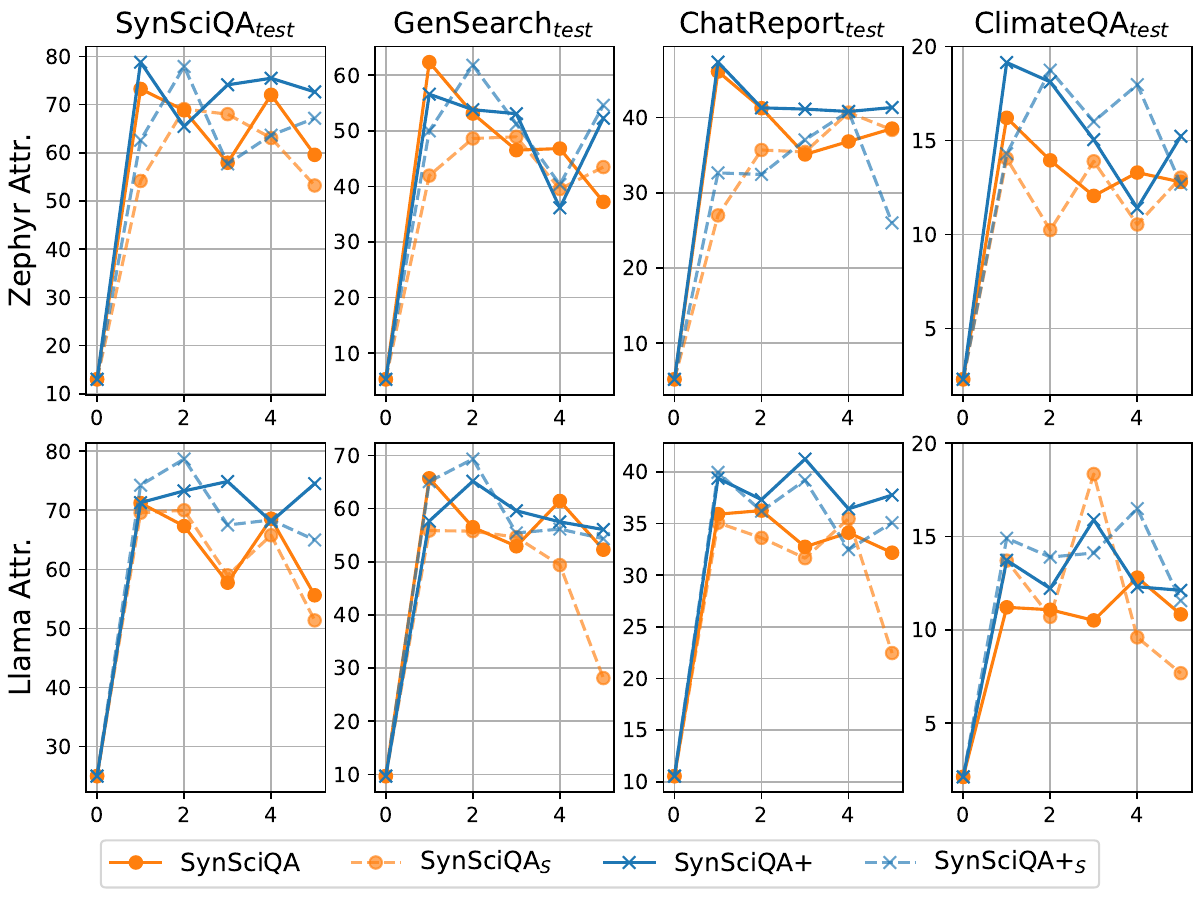}
  \captionsetup{skip=5pt}
	\caption{Controlling \textbf{quality}, Attributability scores vs. number of epoch, caused by different \textbf{quantity}.}
	\label{fig:quantity_attr_all}
% \vspace{-0.7em}
\end{figure}

Given the unsatisfactory performance of open-sourced LLMs on Evidence-Based QA, we want to explore two research questions: \textbf{RQ1.} Do data quality and quantity matter for fine-tuning performance with synthetic data? \textbf{RQ2.} Can synthetic fine-tuning and evaluation contribute to the performance on OOD data and real-world applications? 

To study \textbf{RQ1}, we fine-tune open-sourced LLMs on \textsc{SynSciQA} datasets of different qualities. To control quantity when comparing quality, we randomly sample subsets of \textsc{SynSciQA} and \textsc{SynSciQA+}, leading to \textsc{SynSciQA}$_{S}$ and \textsc{SynSciQA+}$_{S}$ with the same data quantity as \textsc{SynSciQA++}. To study \textbf{RQ2}, we evaluate all fine-tuned checkpoints on test sets of different distributions to see if synthetic data fine-tuning leads to overall improvement. We further calculate the correlation between in-domain (\textsc{SynSciQA}$_{test}$) and OOD (other three test sets) performance to check if in-domain performance can indicate real-world performance. All fine-tuning lasts 5 epochs and we report the performance of all epochs for two reasons: (1) we suspect that epoch number is an essential hyperparameter for OOD performance, as too many epochs may lead to overfitting to synthetic data; and (2) little previous work explores the influence of epoch number and potential overfitting in instructing tuning.

\myparagraph{RQ1: Quality matters more than quantity}. We first compare the fine-tuning performance with data of different quality, having the quantity controlled. \cref{fig:quality_source_all} shows that fine-tuning data with better source quality leads to higher source quality scores (\textsc{SynSciQA+}$_{S}$ and \textsc{SynSciQA++} outperform \textsc{SynSciQA}$_{S}$). \cref{fig:quality_attr_all} shows that higher data quality also leads to better attributability, where in most cases (75\%) \textsc{SynSciQA++} > \textsc{SynSciQA+}$_{S}$ > \textsc{SynSciQA}$_{S}$. Fine-tuning on the highest quality data even leads to comparable or better performance than GPT-4 on \textsc{SynSciQA}$_{test}$ and \textsc{GenSearch}$_{test}$, and GPT-3.5-comparable performance on \textsc{ChatReport}$_{test}$ and \textsc{ClimateQA}$_{test}$.

Furthermore, when we control quality to compare the fine-tuning outcomes of different quantities, we find that more data points do not lead to significant performance improvement, as illustrated in \cref{fig:quantity_source_all} and \cref{fig:quantity_attr_all}. We further conduct statistical tests to verify our observations. The results in \cref{tab:stat_test} show that improving quality leads to statistically significant improvement while only increasing quantity does not. Therefore, we conclude that data quality is more important than quantity for Evidence-Based QA fine-tuning.
\begin{table}[t]
\small
\centering
\begin{tabular}{c@{\hspace{3pt}}c@{\hspace{3pt}}ccc}
\toprule
    \multicolumn{3}{c}{\textbf{Comparison}}& \textbf{Attr.}  & \textbf{Source.}               \\ \hline
\textsc{Syn}$_{S}$ & < & \textsc{Syn}  & 0.3224 & 0.5760 \\
\textsc{Syn+}$_{S}$ & < & \textsc{Syn+}  & 0.8719 & 0.9932 \\
\textsc{Syn}$_{S}$ & < & \textsc{Syn+}$_{S}$  & 2.88e-3$^{*}$ & 7.11e-6$^{**}$ \\
\textsc{Syn+}$_{S}$ & < & \textsc{Syn++}  & 6.13e-5$^{**}$ & 0.1346 \\
\textsc{Syn}$_{S}$ & < & \textsc{Syn++}  & 7.57e-8$^{**}$ & 1.02e-5$^{**}$ \\
\bottomrule
\end{tabular}
\caption{\label{tab:stat_test} Statistical significance of performance difference with different trainsets. We conduct Mann-Whitney U test on different settings and use Fisher's method to merge corresponding p-values (see details in \cref{appendix:stat_test}). By $^{*}$ and $^{**}$, we denote a p-value smaller than $0.01$ and $0.001$, respectively. \textsc{Syn} is short for \textsc{SynSciQA}}
% \vspace{-0.7em}
\end{table}

\myparagraph{RQ2.1: Fine-tuning on synthetic data positively transfers to real world}. It can be observed in \cref{fig:quality_source_all}, \cref{fig:quality_attr_all}, \cref{fig:quantity_source_all}, and \cref{fig:quantity_attr_all} that fine-tuning always lead to better sourcing and attribution performance than original LLMs on in-domain and out-of-distribution test sets. This indicates synthetic data can be used to improve Evidence-Based QA performance in a target domain.

\myparagraph{RQ2.2: Synthetic data as validation set for OOD performance}. We observe a fluctuating performance corresponding to fine-tuning epochs. Therefore, it is important to conduct checkpoint selection over epochs with a validation (or development) set during fine-tuning. However, performance on \textsc{SynSciQA}$_{test}$ is much higher than that of other OOD test sets. Therefore, the performance on a synthetic dataset cannot directly reflect the OOD or real-world performance. But can in-domain synthetic data still be an effective development set indicating which epoch may perform best on OOD data? To answer this question, we compute the Pearson's Correlation between performance scores of different test sets. Results are presented in \cref{tab:corr}, illustrating that the performance on synthetic data has a strong correlation with OOD performance. However, the correlation becomes weaker when the distribution is more distant (\textsc{ClimateQA} and \textsc{ChatReport} has a weaker correlation than \textsc{GenSearch}). Therefore, we conclude that synthetic data can provide a valid development set for OOD performance.

\begin{table}[t]
\small
\centering
\resizebox{\columnwidth}{!}{
\begin{tabular}{c@{\hspace{3pt}}c@{\hspace{3pt}}ccc}
\toprule
    \multicolumn{3}{c}{\textbf{}}& \textbf{Zephyr}  & \textbf{Llama}               \\ \hline
\textsc{SynSciQA}$_{test}$ & \& & \textsc{GenSearch}$_{test}$ Src. & 0.99$^{**}$ & 0.97$^{**}$ \\
\textsc{SynSciQA}$_{test}$ & \& & \textsc{GenSearch}$_{test}$ Attr. & 0.96$^{**}$ & 0.98$^{**}$ \\
\textsc{SynSciQA}$_{test}$ & \& & \textsc{ChatReport}$_{test}$ Attr.  & 0.94$^{**}$ & 0.93$^{**}$ \\
\textsc{SynSciQA}$_{test}$ & \& & \textsc{ClimateQA}$_{test}$ Attr. & 0.91$^{**}$ & 0.94$^{**}$ \\
\bottomrule
\end{tabular}
}
\caption{\label{tab:corr} The performance correlation of different test sets involving all checkpoints of different settings and epochs. The first and second column shows Zephyr-7b-$\beta$ and Llama-2-13b-chat results correspondingly. %By $^{**}$ and $^{***}$, we denote a p-value smaller than $0.001$ and $0.0001$, respectively.
By $^{**}$, we denote a p-value smaller than $0.001$.}
% \vspace{-0.7em}
\end{table}

\begin{figure}[t]
	%\centering
	\includegraphics[width=\columnwidth]{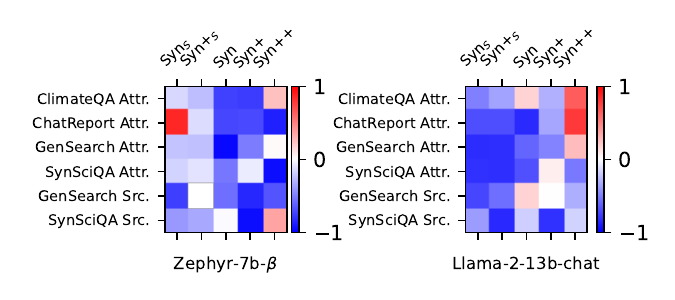}
  \captionsetup{skip=5pt}
	\caption{Correlations between performance and epoch number of all settings.}
	\label{fig:overfit}
% \vspace{-0.3em}
\end{figure}

\myparagraph{RQ2.3: Overfitting does exist}. We suspect that fine-tuning too many epochs may cause overfitting and reduce generalisability. So we compute the Pearson Correlation between the performance scores and epoch numbers of all settings, where positive correlations indicate benefit from more epochs and negative correlations indicate overfits. Results are visualized in \cref{fig:overfit}, showing an overfitting trend for the majority of settings. Therefore, fine-tuning too many epochs may lead to a suboptimal performance. But we do not observe the fine-tuning overfits more to \textsc{SynSciQA} than others. We attribute this to \textsc{SynSciQA}$_{test}$ containing different scientific topics from the training data.

\subsection{Validating Attributability Score}

Although NLI models have been widely applied in previous work for attributability, they might still be prone to make imperfect predictions \citep{yue-etal-2023-automatic}. Therefore, we validate these metrics against human and GPT-4 attributability evaluation. %Once more, we investigate the validity of the attributability score.
Specifically, we randomly sample instruction-answer pairs from various models and all four evaluation benchmarks and ask humans or GPT-4 to annotate whether the \textit{answer attributability} quality dimension holds for each sentence in the answer.

Then, we calculate the Pearson Correlations between NLI-model-based attributability scores and human / GPT-4 evaluated attributability. As \mbox{Table \ref{tab:validationAtt}} shows, the results substantiate the validity of our method for calculating answer attributability. Correlations exceeding 80\% across all comparisons between our scores and those annotated by humans and GPT-4 affirm the mutual reinforcement of the outcomes. For more details, see \cref{appendix:attScoreVal}.

We also notice a potential shortcut to improve the attributability score: a model may only improve its format quality (i.e., providing a citation to more sentences or more correctly writing source names) without improving the answer entailment rate. In \cref{appendix:shortcut}, we provide the entailment ratio of format-correct citations as a side result to show that this shortcut does not exist. Both format and attributability dimensions are improved by fine-tuning.

\begin{table}[t]
\small
\centering
\begin{tabular}{ccc}
\hline
\textbf{\begin{tabular}[c]{@{}c@{}}Human vs.\\ Attributability\end{tabular}} & \textbf{\begin{tabular}[c]{@{}c@{}}GPT-4 vs.\\ Attributability\end{tabular}} & \textbf{\begin{tabular}[c]{@{}c@{}}Human vs.\\ GPT-4\end{tabular}} \\ \hline
0.821**                                                                        & 0.917**                                                                        & 0.871**                                                              \\ \hline
\end{tabular}
\caption{Pearson Correlation between our Attributability Score, Human Annotation, and GPT-4 Annotation. By $^{**}$, we denote a p-value smaller than $0.001$.}
\label{tab:validationAtt}
% \vspace{-1em}
\end{table}

\section{Related Work}
\myparagraph{Basing Answers on Sources}: Prompting LLMs to respond with citations has been a popular pattern of Retrieval Augmented Generation (RAG) for better traceability \citep{karpukhinDensePassageRetrieval2020,lewis2021retrievalaugmented,retro2022,vaghefi2023chatclimate,ni-etal-2023-chatreport,asaiSelfRAGLearningRetrieve2023,liSurveyLargeLanguage2023,saadfalcon2023ares,gaoRetrievalAugmentedGenerationLarge2024}. %, particularly for information-seeking tasks that require accurate knowledge of vertical domain \citep{vaghefi2023chatclimate,semnani2023wikichat,cuiChatLawOpenSourceLegal2023} and high answer traceability \citep{ni-etal-2023-chatreport}. 
However, previous work shows that asking for citations does not make the answer more factually trustworthy \citep{min-etal-2023-factscore}. Commercial search engines and SOTA closed-sourced LLMs suffer from unsatisfactory performance \citep{ni-etal-2023-chatreport,liuEvaluatingVerifiabilityGenerative2023}, while open-sourced LLMs have even worse faithfulness \citep{yue-etal-2023-automatic,gao-etal-2023-enabling}. Therefore, LLMs, especially open-sourced ones, need essential improvement to achieve more trustworthy RAG applications. 

The closest previous study to our work is \citet{gao-etal-2023-enabling} which defines evaluation criteria and benchmarks for citation quality of existing LLMs. However, how to scalably fine-tune open-source LLMs in Evidence-Based QA and rigorously evaluate these specialists in- and out-of-distribution remained an open question.

\myparagraph{Data Distillation for Instruction Tuning}: Distilling instruction-following data from powerful teacher models is an effective and scalable way to improve LLMs' instructing-following performance \citep{honovich-etal-2023-unnatural,wang2023selfinstruct,alpaca,yin-etal-2023-dynosaur,tunstallZephyrDirectDistillation2023}. However, prior research has outlined that simple distillation produces suboptimal data quality \citep{chen2023alpagasus} and that data quality over quantity plays an essential role in improving model output \citep{zhou2023lima}. In this work, we propose that automatic filtering can be a potential way to improve distilled data quality and thus achieve better fine-tuning performance.

\section{Discussions and Future Work}
\myparagraph{Broader Impact}: The aim of this work is to build a basis for constantly improving open-source LLMs in Evidence-Based QA, which is important for the practical community where RAG is heavily employed in applications. The NLP research community may also find our work inspiring in mitigating LLM hallucination: our proposed paradigm for Evidence-Based QA requires all answer sentences to be grounded by in-context sources. Such controlled generation makes hallucination detection much easier leveraging entailment models. Human evaluations in \cref{appendix:handEvalTraining} and \cref{appendix:attScoreVal} also prove the potential of NLI-based hallucination detection.  
\myparagraph{Future Work}: For research, we will continue improving open-sourced LLM's performance on Evidence-Based QA: For example, (1) continuing fine-tuning existing instruction-fine-tuned checkpoints on RLHF alignment stages; (2) generalizing LLM specialists to other templates to study the trade-off between specialization and generalization; and (3) exploring how to leverage LLM parametric knowledge with attributability. For the practical community, we will continuously benchmark new LLMs on our datasets. At the same time, we aim to make the resulting models accessible for the practical community\footnote{See the updates on \url{https://github.com/EdisonNi-hku/Robust_Evidence_Based_QA}.}. Furthermore, we outline that the training data for this project was mainly (75\%) distilled from GPT-3.5 instead of GPT-4, making it more accessible for low-budget RAG development. More powerful generic LLMs for data distillation may improve the results even more.

\section{Conclusion}
In this work, we present a data synthesize pipeline for fine-tuning and evaluating LLMs for Evidence-Based QA. We show that (1) data quality is critical and our quality filters can effectively improve synthetic data quality; (2) synthetic data fine-tuning can improve real-world RAG use case; and (3) synthetic data can make development set indicating OOD performance. Thus, we advocate the view of specializing and focusing LLMs on specific tasks to reach production-ready, real-world applicable solutions.

\section*{Limitations}
As with every work, this study has limitations. First, we only experiment on two open-sourced LLMs: Zephyr-7b-$\beta$ and Llama-2-13b-chat. We hypothesize that the findings are transferable to other pretrained LLMs. We chose this setting because we want to analyze it as comprehensively as possible with our given time and budget restrictions and the broad coverage of investigated aspects including data quantity vs. quantity, out-of-distribution generalisability, and overfitting caused by epoch number. 

Second, due to these budget and time limitations, we also conduct random sampling when performing human and GPT-4 evaluation to verify the attributability score instead of evaluating all instruction-answer pairs in all settings and epochs. However, we argue that the performance of different settings is uniform across different instructions and sampled data points are representative enough. Furthermore, we make all hand evaluations and generations of all checkpoints publicly available (for more details, see \cref{appendix:attScoreVal}).

Third, this work does not fully assess the quality dimension of helpfulness. We seek to improve open-sourced LLMs in the dimensions of faithfulness and answer-traceability, the most significant shortcomings of open-sourced models. We argue that helpfulness is hard to define and leave it to future exploration (see \cref{appendix:helpfulnessEval}).

Fourth, this work only explores a single prompt template for Evidence-Based QA, which states our quality dimensions and extra requirements for better traceability (e.g., one sentence one citation). Since we conduct no prompt engineering/optimization, we hypothesize that the core findings of this work are transferable to other use cases where prompt templates need to be different (e.g., different citation format, evidence-grounded RAG tasks other than QA). Specifically, practitioners may depend on their own need to write prompt templates and define quality filters to improve distilled data. We plan to verify this in future work.

\section*{Ethics Statement}
\myparagraph{Human Annotation}: In this work, all human annotators are Doctorate, Post-Doc researchers or Professors who have good knowledge about scientific communication and entailment. They are officially hired and have full knowledge of the context and utility of the collected data. We adhered strictly to ethical guidelines, respecting the dignity, rights, safety, and well-being of all participants. 

\myparagraph{Data Privacy or Bias}: There are no data privacy issues or bias against certain demographics with regard to the data collected from real-world applications and LLM generations. All artifacts we use are under a creative common license. We also notice no ethical risks associated with this work.

\myparagraph{Reproducibility Statement}: To ensure full reproducibility, we will disclose all codes and data used in this project, as well as the LLM generations, GPT-4 and human annotations. For OpenAI models, we use gpt-3.5-turbo-0613 and gpt-4-0613 for synthetic data generation and gpt-4-turbo-0125-preview for GPT-4 evaluation (due to the project timeline, we do not use gpt-4-turbo-0125-preview for synthetic data generation). We always fix the temperature to 0 when using APIs.

\ifarxiv
\section*{Acknowledgements} 
% Hi Markus. Für die finale Abgabe sind die Acknowledgements drin.
This paper has received funding from the Swiss
National Science Foundation (SNSF) under the project `How sustainable is sustainable finance? Impact evaluation and automated greenwashing detection' (Grant Agreement No. 100018\_207800).  It is also funded by grant from Hasler Stiftung for the Research Program Responsible AI with the project ``Scientific Claim Verification.''
\else

\fi

\bibliography{custom}

\appendix

\section{The Challenge of Helpfulness} \label{appendix:helpfulnessEval}

\begin{table*}[ht]
\small
\centering
\resizebox{\textwidth}{!}{
\begin{tabular}{lcccccccccccc}
\toprule
& \multicolumn{3}{c}{\textbf{\textsc{SynSciQA}$_{test}$}} & \multicolumn{3}{c}{\textbf{\textsc{GenSearch}$_{test}$}} & \multicolumn{3}{c} {\textbf{\textsc{ChatReport}$_{test}$}} & \multicolumn{3}{c} {\textbf{\textsc{ClimateQA}$_{test}$}} \\
\cmidrule(lr){2-4} \cmidrule(lr){5-7} \cmidrule(lr){8-10} \cmidrule(lr){11-13}  
& \begin{tabular}[c]{@{}c@{}}avg. \#\\ sentences\end{tabular} & \begin{tabular}[c]{@{}c@{}}avg. length\\ sentences\end{tabular} & \begin{tabular}[c]{@{}c@{}}avg. unique \\ citations\end{tabular} & \begin{tabular}[c]{@{}c@{}}avg. \#\\ sentences\end{tabular} & \begin{tabular}[c]{@{}c@{}}avg. length\\ sentences\end{tabular} & \begin{tabular}[c]{@{}c@{}}avg. unique \\ citations\end{tabular} & \begin{tabular}[c]{@{}c@{}}avg. \#\\ sentences\end{tabular} & \begin{tabular}[c]{@{}c@{}}avg. length\\ sentences\end{tabular} & \begin{tabular}[c]{@{}c@{}}avg. unique \\ citations\end{tabular} & \begin{tabular}[c]{@{}c@{}}avg. \#\\ sentences\end{tabular} & \begin{tabular}[c]{@{}c@{}}avg. length\\ sentences\end{tabular} & \begin{tabular}[c]{@{}c@{}}avg. unique \\ citations\end{tabular} \\
\midrule
Llama-\textsc{Syn} & $\underline{4.30}$ & $27.01$ & $\underline{2.01}$ & $\underline{2.76}$ & $20.82$ & $\underline{1.31}$ & $\underline{4.21}$ & $24.44$ & $\underline{1.67}$ & $\underline{5.34}$ & $24.24$ & $\underline{3.15}$ \\
Llama-\textsc{Syn++} & $3.89$ & $\underline{28.06}$ & $1.83$ & $2.13$ & $\underline{21.26}$ & $1.25$ & $3.78$ & $\underline{25.73}$ & $1.46$ & $4.15$ & $\underline{24.71}$ & $2.88$ \\ \midrule
Zephyr-\textsc{Syn} & $\underline{4.37}$ & $\underline{27.72}$ & $\underline{2.15}$ & $\underline{2.62}$ & $\underline{19.56}$ & $\underline{1.38}$ & $\underline{4.58}$ & $\underline{24.97}$ & $\underline{1.94}$ & $\underline{5.20}$ & $\underline{25.24}$ & $\underline{3.49}$ \\
Zephyr-\textsc{Syn++} & $3.89$ & $26.50$ & $1.56$ & $2.08$ & $19.04$ & $1.27$ & $3.80$ & $23.60$ & $1.56$ & $4.26$ & $22.73$ & $2.70$ \\
\bottomrule
\end{tabular}
}
\caption{Statistics of fine-tuned models' outputs. Larger values are \underline{underlined}. The table reports the second-epoch checkpoints' outcomes.}
\label{tab:output_stat}
% \vspace{-0.7em}
\end{table*}
Helpfulness can be defined as ''How well does the answer address the question?''. We argue that helpfulness is extremely hard to evaluate in Evidence-Based QA.

Following the definition of helpfulness, one could argue that the comparison of question-answer pairs can yield insights into helpfulness. If the model answers the question well, then it is helpful. However, as Table \ref{tab:example1helpfulness} shows, this undermines the logic of answering based on sources in Evidence-Based QA. If no sources are given, then the answer should reflect that. Following the definition, this might be less helpful but certainly more faithful. We argue that this example rather shows that helpfulness and faithfulness are intertwined. Therefore, we view that our source quality score partially addresses helpfulness by indicating whether the answer is only based on valid sources.

Secondly, generations of fine-tuned models are driven by the distribution fine-tuning data. As illustrated in \cref{tab:output_stat}, models fine-tuned on \textsc{SynSciQA++} result in slightly shorter answers and a smaller number of unique citations than \textsc{SynSciQA}. This perfectly reflects the training data distribution (see \cref{tbl:trainDataStatistics}). Following the definition of helpfulness, one could argue that more context and therefore more answer length is more helpful. However, longer answers with more citations do not indicate more helpfulness in Evidence-Based QA. Table \ref{tab:example2helpfulness} shows an example where one answer sentence - irrespective of source quality - concisely answers the question while the other provides extra context. Is the answer with more context more helpful? We argue that it highly depends and is therefore not easily evaluable. However, through the lens of Evidence-Based QA, an answer is only helpful if the cited sources entail the answer. Thus, also our second metric of answer attributability partially addresses helpfulness. In addition, if more lengthy answers are preferred, one can easily achieve that by encouraging longer answers when generating and filtering synthetic data, as shown by \cref{tab:output_stat} and \cref{tbl:trainDataStatistics}. 

Collectively, we view the exact measurement of helpfulness as a challenge for future work. However, we argue that our two employed evaluation metrics already address helpfulness in Evidence-Based QA to a satisfactory degree. 

To improve the helpfulness evaluation, future work could try to identify dimensions of helpfulness that are perpendicular to source quality and answer attributability and evaluate them with the help of LLMs. One dimension could be friendliness or the degree how well the question is addressed. However, as outlined, these dimensions might stand in conflict with the two quality dimensions introduced in this work. Thus, investigating these trade-offs could present an interesting new direction.

\begin{table*}[t]
\small
\centering
\resizebox{\textwidth}{!}{
\begin{tabular}{>{\raggedright\arraybackslash}m{0.25\textwidth}m{0.75\textwidth}}
\hline
\textbf{Question:} & \multicolumn{1}{c}{\textbf{How does natural selection interact with genetic drift?}} \\ \hline

Answer by GPT-4 & \textit{None of the provided sources address the interaction between natural selection and genetic drift.}
 \\ \hline
 
Answer by Llama-13b-chat & \textit{Sure, I'd be happy to help! Natural selection and genetic drift are both important mechanisms that drive the evolution of species. While natural selection is the process by which individuals with advantageous traits are more likely to survive and reproduce, genetic drift is the random change in the frequency of alleles in a population over time \colorbox{goldenrod}{(Morse et al., 2018, p. 67)}. The interaction between natural selection and genetic drift can be complex, as both processes can influence the evolution of a species in different ways. For example, natural selection can lead to the fixation of advantageous alleles in a population, while genetic drift can result in the loss or fixation of neutral or deleterious alleles \colorbox{goldenrod}{(Hudsonsonian Institution, 2017, p. 28)}. Additionally, \textbf{[...]}}
\\ \hline

Answer by Llama-13b-chat fine-tuned for two epochs on \textsc{SynSciQA++} & \textit{None of the provided sources address the question "How does natural selection interact with genetic drift?" Therefore, an answer cannot be given based on these sources.}
\\ \hline
\end{tabular}
}
\caption{\label{tab:example1helpfulness}Answers to the same prompt containing no question-relevant sources by GPT-4, plain vanilla LLama-13b-chat and LLama-13b-chat fine-tuned for two epochs on the high-quality dataset \textsc{SynSciQA++}. Cited relevant sources are marked in \colorbox{green}{green}, and irrelevant or erroneous sources are marked in \colorbox{goldenrod}{brown}.}
% \vspace{-0.7em}
\end{table*}

\begin{table*}[t]
\small
\centering
\resizebox{\textwidth}{!}{
\begin{tabular}{>{\raggedright\arraybackslash}m{0.25\textwidth}m{0.75\textwidth}}
\hline
\textbf{Question:} & \multicolumn{1}{c}{\textbf{Who is the top ranked female tennis player in the world?}} \\ \hline

Answer by GPT-4 & \textit{The current top-ranked female tennis player in the world is Iga Swiatek \colorbox{green}{(Online2602022, 2019, p.8)}.}
 \\ \hline
 
Answer by Zephyr-7b-$\beta$ & \textit{As of the sources provided, Iga Swiatek currently holds the top ranking in women's singles tennis. This information can be found in the source "Online2602022" \colorbox{goldenrod}{(Williams, 2022, p.8)}. Ashleigh Barty, the previous world No.1, retired early in 2022, which led to Swiatek's ascension to the top spot. \colorbox{goldenrod}{(Williams, 2022, p.8)}}
\\ \hline

Answer by Zephyr-7b-$\beta$ fine-tuned for two epochs on \textsc{SynSciQA++} & \textit{The top ranked female tennis player in the world is Iga Swiatek \colorbox{green}{(Online2602022, 2019, p.8)}.}
\\ \hline
\end{tabular}
}
\caption{\label{tab:example2helpfulness}Answers to the same prompt containing question-relevant sources by GPT-4, plain vanilla Zephyr-7b-$\beta$ and Zephyr-7b-$\beta$ fine-tuned for two epochs on the high-quality dataset \textsc{SynSciQA++}. Cited relevant sources are marked in \colorbox{green}{green}, and irrelevant or erroneous sources are marked in \colorbox{goldenrod}{brown}.}
% \vspace{-0.7em}
\end{table*}

\section{Attributability Score Details} \label{appendix:faithfulnessDef}
We use SpaCy \citep{spacy2} to split answers into sentences. Unattributable answer sentences caused by missing citations or wrong citation format can be easily identified through golden heuristics, for example, matching citations with actual source names. However, it is hard for a heuristic-based method to judge whether a statement is entailed by the cited source. Previous work proposes to use NLI models to predict entailment \citep{honovich-etal-2022-true-evaluating,gao-etal-2023-enabling,yue-etal-2023-automatic}. Among them, \citet{yue-etal-2023-automatic} aggregate the largest NLI training set and conduct extensive analyses to explore the best practice of attributability prediction. Therefore, we rely on their results to select models for the attributability score. The two best-performing checkpoints are Flan-t5-XL and Flan-t5-XXL \footnote{https://huggingface.co/osunlp/attrscore-flan-t5-xxl}. When inferencing with these checkpoints, we follow the prompt template in \citet{yue-etal-2023-automatic} and use greedy generation. We aggregate the prediction of both models to improve the precision since false positives are more harmful than false negatives in the task of judging LLM faithfulness.

The design of our attributability score mostly follows the citation recall score of \citet{gao-etal-2023-enabling}. However, we only calculate attributability scores on answers with at least one citation, which differs from \citet{gao-etal-2023-enabling}, because we also consider scenarios where there is no relevant source at all. In that case, the model should state no source is relevant without any citation.

\section{Training Data Creation Process} \label{appendix:dataCreation}
For creating the raw data, we employ several steps. For all creation steps, we use the June checkpoints of GPT-3.5-turbo and GPT-4. In the following, the used prompts are displayed. First, we create a set of 100+ random topics with the help of GPT-4 using the following prompt.
\begin{lstlisting}[frame=single, basicstyle=\ttfamily\scriptsize, xleftmargin=0pt, breaklines, numbers=none, xleftmargin=.05\columnwidth, xrightmargin=.05\columnwidth]
"Create {n} random topics from the scientific areas of finance, sustainability, physics, social sciences and natural sciences. Please seperate each topic with '||'. Use no enumeration or additional signs to seperate the topics."
\end{lstlisting}

This results in a broad span of topics ranging from \textit{"Corporate finance"} over \textit{"Anthropology"} and \textit{"Electromagnetism"} to \textit{"Dark matter"}. Following this first step, we create 25 questions per topic with GPT-4 (see below).

\begin{lstlisting}[frame=single, basicstyle=\ttfamily\scriptsize, xleftmargin=0pt, breaklines, numbers=none, xleftmargin=.05\columnwidth, xrightmargin=.05\columnwidth]
"Take the topic {topic} and create {n} questions that could be posed in the field. Make the questions diverse and differentiable from each other.

End every question with '\\'. Use no enumeration or additional signs to seperate the questions."
\end{lstlisting}

Furthermore, we create three paragraphs that address the question as an artificial source with both GPT-3.5 and GPT-4. A random variable is introduced that enforces the creation of around 25\% of the data points with GPT-4 to enhance the diversity of the training dataset distribution. The exact final percentage for data created with GPT-4 is 24.97\%.

\begin{lstlisting}[frame=single, basicstyle=\ttfamily\scriptsize, xleftmargin=0pt, breaklines, numbers=none, xleftmargin=.05\columnwidth, xrightmargin=.05\columnwidth]
"Consider the following question within the topic {topic}: {question}

Please create {m} paragraphs with the length of 2-4 sentences that partially address this question. The question should not fully be answered by one paragraph but rather helpful content in respect to the question should be displayed. Each paragraph should be in the style of a book or research article.

Furthermore, the paragraphs can display different perspectives and should not overlap much. The paragaphs should also alternate in level of detail and addressed readers, i.e., some paragraphs can be very scientifc while others would rather serve a general public.

It is important that the paragraphs stand for themselves. They don't read like one article but excerpts from multiple articles.

Please be creative with the beginning of the paragraphs.

In the end of each paragraph give author, year and page in the following format '[author, year, page]'. Follow this example: '[Mishra et al., 2019, p.54]'.

Make up author, year and page, if you don't have this information. Authors can also be institutions.

End every paragaph with 'ENDOFPARAGRAPH'. Use no enumeration or additional signs to seperate the paragraphs. Also do not give any further information like "Paragraph 1: ...".
\end{lstlisting}

Finally, we design an instruction that contains 0-3 \textit{relevant} sources that stem from the paragraphs created above, and 3-6 \textit{irrelevant} sources that do not correspond to the question (for a template, see Prompt Template in \cref{sec:22}). For selecting the \textit{irrelevant} sources, we randomly sample sources from other topics in the dataset. We use GPT-3.5-turbo and GPT-4 to create an answer according to the source creation. This results in the \textsc{SynSciQA} dataset.

Finally, we apply the source quality filter to obtain \textsc{SynSciQA+} and the answer attributability filter to obtain \textsc{SynSciQA++}. \cref{tbl:trainDataStatistics} shows that the instructions stay comparatively similar throughout the filtering process. For the answers, the number of unique citations and the average sentence number slightly decreases after applying the source quality filter and the attributability quality filter correspondingly, indicating that these filters may effectively filter out answers with problematic citations and unattributable statements. This likely coincides with a higher probability of short paragraphs containing fewer errors. However, both mechanisms don't seem to largely influence answer length and number of cited sources. Rather, the intended behavior of concise answers might be strengthened.

\begin{table*}[t]
\small
\centering
\begin{tabular}{clcccclccc}
\toprule
           &  & \multicolumn{4}{c}{Instructions}                                                                                                                      &  & \multicolumn{3}{c}{Answers}                                                                                                                                                                      \\ \cline{1-1} \cline{3-6} \cline{8-10} 
Dataset    &  & \# samples & avg. \# srcs & \begin{tabular}[c]{@{}c@{}}\# words/src\\ avg.\end{tabular} & \begin{tabular}[c]{@{}c@{}}\# words/src\\ std.\end{tabular} &  & \begin{tabular}[c]{@{}c@{}}avg. \#\\ sentences\end{tabular} & \begin{tabular}[c]{@{}c@{}}avg. length\\ sentences\end{tabular} & \begin{tabular}[c]{@{}c@{}}avg. unqiue \\ citations\end{tabular} \\ \cline{1-1} \cline{3-6} \cline{8-10} 
\textsc{SynSciQA}   &  & 2143       & 6.11         & 112.08                                                      & 21.64                                                       &  &                     4.28                                        &        27.22                                                         &              1.93                                                    \\
\textsc{SynSciQA+}  &  & 1386       & 6.27         & 112.21                                                      & 21.57                                                       &  &                                 4.20                            &         26.78                                                        &          1.65                                                        \\
\textsc{SynSciQA++} &  & 669        & 6.09         & 112.39                                                      & 21.41                                                       &  &          3.87                                                   &             28.49                                                    &         1.62                                                         \\ \bottomrule
\end{tabular}
\caption{Overview of the statistical characteristics of the training datasets.}
\label{tbl:trainDataStatistics}
\end{table*}

\section{Hand-Evaluation of the Quality Filters} \label{appendix:handEvalTraining}
The hand-evaluation of the \textsc{SynSciQA++} dataset centers around the entailment quality, i.e. whether the answer is entailed by the source. The other two quality dimensions, source, and format, can be controlled automatically, i.e. there are only the right sources in the answers and each sentence ends with a source. To control the entailment quality, we randomly sample 300 source-answer pairs that are evaluated by two annotators. The two annotators per sample stem from four researchers including two doctorate researchers, one post-doctorate researcher, and one professor. As \cref{tbl:handEval_Entailment} shows, the overwhelming amount of answers is correctly entailed by the source.

\begin{table}[t]
\centering
\begin{tabular}{lc}
\hline
\textbf{}                                                                               & \textbf{Percentage} \\ \toprule
\begin{tabular}[l]{@{}l@{}}Both annotators agree on\\ correct entailment\end{tabular}   & 94.3\%              \\ \hline
\begin{tabular}[l]{@{}l@{}}Both annotators agree on\\ incorrect entailment\end{tabular} & 2.0\%               \\ \hline
Split decisions                                                                         & 3.7\%               \\ \bottomrule
\end{tabular}
\caption{Hand-Evaluation Results on Entailment Quality.}
\label{tbl:handEval_Entailment}
\end{table}

On the one hand, only 2\% of the data is not rightfully entailed. These mainly originate from samples where the model replicated the details in a slightly wrong manner. One example can be seen in \cref{lst:incorrectEntailment}. The answer states that the main organs of the digestive system are the mouth, esophagus, stomach, small intestine, and large intestine. However, this is only one part of the answer. The main organs also comprise the accessory organs (see \cref{lst:incorrectEntailment}).

\begin{figure*}[ht]
\begin{lstlisting}[frame=single, basicstyle=\ttfamily\scriptsize, xleftmargin=0pt, numbers=none]
SOURCE:  The digestive system is a fascinating and intricate system that allows our bodies to obtain the necessary nutrients for survival. While many organs play a role in this process, the main organs of the digestive system can be categorized into two groups: the gastrointestinal tract and the accessory organs. The gastrointestinal tract consists of the mouth, esophagus, stomach, small intestine, and large intestine, which work together to break down food and absorb nutrients. The accessory organs, on the other hand, include the liver, gallbladder, and pancreas, which produce and release substances that aid in digestion. By understanding the main organs of the digestive system, we can appreciate the complexity of this system and the importance of maintaining its health. 

ANSWER SENTENCE: [...] The main organs of the digestive system are the mouth, esophagus, stomach, small intestine, and large intestine. [...] 

HUMAN VERDICT: Two times incorrect entailment. The main organs of the digestive system also comprise the accessory organs.
\end{lstlisting}
\caption{\label{lst:incorrectEntailment} Example for an incorrect entailment.}
\end{figure*}

On the other hand, 4\% of the cases were split decisions. These predominately originate from different interpretations of nuances in the used language. Disagreements are resolved through debating about specific meanings of nuances until a concensus is achieved. \cref{lst:partlyIncorrectEntailment} shows an example of this. 

\begin{figure*}[ht]
\begin{lstlisting}[frame=single, basicstyle=\ttfamily\scriptsize, xleftmargin=0pt, numbers=none ]
SOURCE: From a more technical perspective, working capital management involves the optimization of a firm's liquidity position by managing the trade-offs between profitability and risk. This is achieved by managing the components of working capital, namely accounts receivable, inventory, and accounts payable. The management of these components involves determining the optimal level of investment in each, considering the costs and benefits associated with different levels of investment. For instance, while a high level of inventory may reduce the risk of stock-outs, it also ties up funds that could be used elsewhere in the business.

ANSWER SENTENCE: [...] This is achieved by managing the components of working capital, which include accounts receivable, inventory, and accounts payable. [...] 

HUMAN VERDICT: One time incorrect entailment. The word "include" means that the following objects are necessary but not sufficient while "namely" in the source signals that they are sufficient and necessary.
\end{lstlisting}
\caption{\label{lst:partlyIncorrectEntailment} Example for an incorrect entailment.}
\end{figure*}

This analysis shows the limits of the automatic filters that can deal with a good amount of cases but fail to detect the last bit of small nuances. However, since the vast majority of pairs are valid, the quality filters seem to perform the intended way.

\section{Creation of \textsc{GenSearch}$_{test}$} \label{appendix:EvalBenchGenSearch}
\textsc{GenSearchEngines-test} is developed from the dataset created by \citet{liu2023evaluating}. In this project, the authors create a dataset from generative search engines such as Bing Chat, perplexity AI, or NeevaAI. The task in the project is to hand-evaluate different quality dimensions of the answers. Thus, the annotators are presented with queries of these tools and investigate the given sources. Amongst others, they answer whether the source is accurate in answering the question. 

We make use of this dataset and hand-check 600 question-source pairs. While evaluating, we quickly identify that some questions should not be taken into account because they are inconclusive, vague, or impractical for other reasons. For instance, the dataset contains questions like \textit{"tips to win fight at school"} or \textit{"Deep web?"}. Additionally, not all sources were practical or necessary to respond to a question. Some questions contained more than ten sources and others contained duplicates. Thus, we hand-filtered 276 question-source pairs that we deemed relevant. This resulted in 106 unique questions with an average of 2.6 sources. We further processed incomplete questions into a question form. For instance, we added a question mark to each question and added fill words if needed to properly understand the question.

Since we now again have a dataset that contains \textit{relevant} and \textit{irrelevant} sources, we can reiterate the steps used for creating \textsc{SynSciQA} (see Step 4 in \cref{sect31}). This way, we create a dataset that is similar in structure but different in the underlying distribution of the data sources and questions. First, the questions are now rather practical and not scientific anymore. This also translates to the source space that is now rather from websites or online blogs. Furthermore, the sources do not necessarily contain full sentences and are usually written by humans in simple language (assuming that online articles are written by humans). Since the topic range is much more diverse, the resulting instructions usually contain sources where a human evaluator could clearly state which sources belong to the question. This combination of simple language and a very wide range of questions theoretically makes the differentiation between \textit{relevant} and \textit{irrelevant} sources much easier.

% \section{Aligned Model vs. Unaligned Model} \label{appendix:alignedornot}

\section{Test Data Examples} \label{appendix:test_example}
To outline the differences between the test datasets, we further explore the properties of each evaluation benchmark. It is important to outline that we gradually leave distance from the properties of the in-domain dataset. This way, we ultimately aim to obtain insights into the real-world applicability of the approach.

The first evident difference lies in the statistical properties of the sources in the instructions (see Table \ref{tbl:statisticsTestDatasets}). We create \textsc{SynSciQA}$_{test}$ and \textsc{GenSearch}$_{test}$ with the data creation pipeline described in \cref{sec:22}. This means, we have known \textit{relevant} and \textit{irrelevant} sources in the datasets. On the other hand, we use top-10 retrieved sources for \textsc{ChatReport}$_{test}$ and the output sources by ClimateQA for \textsc{ClimateQA}$_{test}$. As Table \ref{tbl:statisticsTestDatasets} shows, the different sourcing mechanisms result in a large difference in average length and standard deviation. 

Furthermore, there are large differences in the structure and format of the sources. The exemplary comparison in Table \ref{tab:datasetSourcesExamples} reveals that \textsc{SynSciQA}$_{test}$ and \textsc{GenSearch}$_{test}$ are predominantly in full-sentence form. While \textsc{SynSciQA}$_{test}$ only contains synthetic scientific topics, \textsc{GenSearch}$_{test}$ is created from internet sources and therefore much broader and more colloquial in tone. Sources in \textsc{ChatReport}$_{test}$ start in the middle of the sentence, end in the middle of the sentence, and are not necessarily in full-sentence form. The same holds true for \textsc{ClimateQA}$_{test}$. However, this dataset also contains nested citations which represents the most complicated case for Evidence-Based QA.

\begin{table}[t]
\small
\centering
\begin{tabular}{ccccc}
\toprule
\multicolumn{1}{l}{} & \multicolumn{1}{l}{}                                  & \multicolumn{1}{l}{}                                    & \multicolumn{2}{c}{\# words per src} \\
Dataset              & \begin{tabular}[c]{@{}c@{}}\# \\ samples\end{tabular} & \begin{tabular}[c]{@{}c@{}}avg. \# \\ srcs\end{tabular} & avg.              & std.             \\ \hline
\textsc{SynSciQA}$_{test}$      & 539                                                   & 6.10                                                     & 98.23              & 21.28             \\
\textsc{GenSearch}$_{test}$      & 106                                                   & 5.85                                                     & 82.31              & 41.80             \\
\textsc{ChatReport}$_{test}$     & 110                                                   & 10.00                                                      & 67.94              & 25.27             \\
\textsc{ClimateQA}$_{test}$      & 261                                                   & 4.47                                                     & 134.51             & 13.35             \\ \bottomrule
\end{tabular}
\caption{Number of instructions per dataset, average sources per instruction as well as average and standard deviation of number of words per instruction for the test datasets.}
\label{tbl:statisticsTestDatasets}
\end{table}

\begin{table*}[t]
\small
\centering
\resizebox{\textwidth}{!}{
\begin{tabular}{>{\raggedright\arraybackslash}m{0.15\textwidth}m{0.85\textwidth}}
\hline
\textbf{Dataset} & \multicolumn{1}{c}{\textbf{Example Source Paragraph}} \\ \hline

\textsc{SynSciQA}$_{test}$ & \textit{The electromagnetic force is responsible for the most familiar interactions in our everyday lives. It is the force that allows us to see, feel, and interact with the world around us. When light interacts with matter, it can be absorbed, reflected, or transmitted, giving rise to the colors we perceive. The electromagnetic force also enables the operation of electronic devices, such as computers and smartphones, by allowing the flow of electric currents. Moreover, this force is essential for generating and transmitting electrical power. Understanding how particles interact through the electromagnetic force is not only of interest to scientists but also has practical applications that impact our daily lives.}
 \\ \hline
 
\textsc{GenSearch}$_{test}$ & \textit{But how?! Well, there’s merchandising, VOD, streaming video, foreign sales, and a plethora of other distribution channels that can help filmmakers, producers, and studios turn a profit. Traditionally, movies have made their money from ticket sales at the box office or in theaters. A studio might make about 60\% of a film’s ticket sales in the United States, and around 20\% to 40\% of that on overseas ticket sales.}
\\ \hline

\textsc{ChatReport}$_{test}$ & \textit{the details of Fortive’s climate-related governance, strategy, risk management, and metrics and targets. In 2021, we formally expanded the risk criteria within our Enterprise Risk Management program to account for the financial, operational, and regulatory risks in addition to physical risks for which we were already accounting. By incorporating additional climate-related risks into our existing protocol for evaluating and identifying risk, we are able to capture climate-related}
\\ \hline

\textsc{ClimateQA}$_{test}$ & \textit{5 that are subducted over decades are expected to experience signiﬁcant warming (see Figure 5.3). The warming in the subtropical gyres penetrates deeper into the ocean than other gyres (roughly 15ºN–45ºN and 15ºS–45ºS in Figure 5.3), following the wind-driven bowing down of the density surfaces (the solid lines in Figure 5.3) in these gyres (Terada and Minobe, 2018). The greater warming at 700–2000 m in the Atlantic than the Paciﬁc or Indian Oceans (Figure 5.3) reﬂects the strong southward transport of recently formed NADW at these depths by the AMOC. Two areas that commonly exhibit substantially reduced near-surface warming over the course of the 21st century are the northern north Atlantic, where a slowing AMOC (see Section 6.7.1.1) reduces the northward heat transport and brings the surface temperatures closer to what is found in other ocean basins at these latitudes (Collins et  al. 2013), and the southern side of the Southern Ocean, where water upwells}
\\ \hline
\end{tabular}
}
\caption{\label{tab:datasetSourcesExamples}Examples of source paragraphs for the different (training/testing) datasets. The structure of \textsc{SynSciQA}$_{test}$ is representative of \textsc{SynSciQA}.}
% \vspace{-0.7em}
\end{table*}

\section{Hyperparameter and Other Settings} \label{appendix:hyperparameter}
We always use random seed 42 for experiments in this work. We use the default QLoRA hyperparameter settings \footnote{https://github.com/jondurbin/qlora}, namely, an effective batch size of 32, a lora r of 64, a lora alpha of 16, a warmup ratio of 0.03, a constant learning rate scheduler, a learning rate of 0.0002, an Adam beta2 of 0.999, a max gradient norm of 0.3, a LoRA dropout of 0.1, 0 weight decay, a source max length of 2048, and a target max length of 512. We use LoRA module on all linear layers. 

We always use SpaCy for word count and sentence split, and Scipy to compute Pearson's Correlation and other statistical significance tests.

All experiments are conducted on two clusters, one with 4 V100 GPUs and the other with 4 A100 (80G) GPUs. 1 GPU hour is used per fine-tuning.

This hyperparameter setup of training epochs orientates on previous impactful and practical work in the domain.\footnote{e.g., training for 2 epochs in \url{https://aclanthology.org/2023.emnlp-main.245/} or for 3-5 epochs in \url{https://github.com/tatsu-lab/stanford_alpaca} or \url{https://magazine.sebastianraschka.com/p/practical-tips-for-finetuning-llms} that argue that multi-epoch training does not benefit LoRA.} However, extending the study 5 to 10 or 15 epochs would likely make some arguments stronger.

\section{Relevance Label for Real RAG} \label{appendix:relevance_label}
\cref{sec:eval_sets} introduces that source-relevance label is available for \textsc{SynSciQA}$_{test}$ thanks to the data creation process. We also annotate source-relevance for \textsc{GenSearch}$_{test}$. However, we do not annotate that for \textsc{ChatReport}$_{test}$ and \textsc{ClimateQA}$_{test}$ because we find most of the retrieved top-k sources in those real RAG systems are directly or indirectly relevant since they are retrieved from a narrow domain (e.g., a sustainability report). Take the following source-question pair as an example:

\begin{itemize}
  \item Question: How resilient is the organisation's strategy when considering different climate-related scenarios, including a 2°C target or lower scenario? How resilient is the organisation's strategy when considering climate physical risks? 
  \item Source: ... Risk Management a. Describe the organization's processes for identifying and  assessing climate-related risks. CDP C2.1 CDP C2.2 CDP C2.2a Risk Management b. Describe the organization's processes for managing  climate-related risks. CDP C2.1 CDP C2.2 Risk Management c.  Describe how processes for identifying, assessing, and managing  climate-related risks are integrated into the organization's overall  risk management. CDP C2.1 CDP C2.2 Metrics and Targets ...
\end{itemize}

Although the source does not directly address the resilience of the company's strategy considering climate risks, it provides information about the company's climate-related risk management, which can be indirectly useful for the resilience considering climate-related risk. Therefore, we rely on answer attributability to evaluate the real-world RAG test sets. As long as the answers have good traceability, we assume relevant information is provided to the question.
\section{Statistical Significance Tests} \label{appendix:stat_test}
To show the statistical significance of performance difference in \cref{sec:fine-tuning}, we first conduct Mann-Whitney U test on each sub-figure of \cref{fig:quality_source_all}, \cref{fig:quality_attr_all}, \cref{fig:quantity_source_all}, and \cref{fig:quantity_attr_all}. Specifically, we regard the scores of epoch 1 to 5 comes from the same distribution, and compute if distributions of different settings are statistically significantly different or not. For example, \textsc{SynSciQA++} distribtuion ([81.56, 81.59, 80.83, 78.19, 81.9]) and \textsc{SynSciQA}$_{S}$ distribtuion ([48.15, 62.01, 61.17, 57.05, 52.57]) in the first sub-figure of \cref{fig:quality_attr_all}. We use Mann-Whitney U test instead of student-t test to avoid making the normal distribution assumption. After having p-values between all settings, we apply Fisher's method to aggregate the p-values, resulting in \cref{tab:stat_test}.

\section{Validating the Attributability Score} \label{appendix:attScoreVal}
The target of this validation is to reinforce the validity of the employed methodology in the \textit{answer attributability}. Generally, this investigation follows the same structure as our train set validation in \cref{appendix:handEvalTraining}. The main difference is that we now investigate all settings, including those out-of-distribution and using open-source models to find evidence that the comparisons are valid. To investigate the decisions, we repeat the evaluation of our score with both human and GPT-4 annotation. Both evaluations follow the structure of \textit{answer attributability} and are articulated through the following prompt.

\begin{lstlisting}[frame=single, basicstyle=\ttfamily\scriptsize, xleftmargin=0pt, breaklines, numbers=none, xleftmargin=.05\columnwidth, xrightmargin=.05\columnwidth]
"Your task is to evaluate whether a SENTENCE represents the information in a SOURCE. This criterion is defined as faithfulness. Faithfulness answers the main question of "Is the SENTENCE content justified through the SOURCE?". The SENTENCE should reflect the information given in the SOURCE. If the SOURCE information does not entail the SENTENCE, then the SENTENCE is not faithful. The SENTENCE must not contain completely new details that are not mentioned in the SOURCE. However, if the SENTENCE contains the same meaning as the SOURCE but only the wording changes, the SENTENCE is still faithful.

SOURCE: +++ {0} +++

SENTENCE: ||| {1} |||

Answer whether the ANSWER is faithful with respect to the SOURCE given the above definition of faithfulness. Respond by starting with "[[YES]]" or "[[NO]]" and then justify your decision in at most one sentence.
"
\end{lstlisting}

For human evaluation, we evaluate 8 settings in total: raw models include GPT-3.5, GPT-4, Llama-2-13b-chat, and Zephyr-7b-$\beta$; fine-tuned models include Llama-2-13b-chat and Zephyr-7b-$\beta$ trained on \textsc{SynSciQA} and \textsc{SynSciQA++} for 2 epochs. We choose the second epoch since it usually does not associate with strong over- or under-fitting. For each setting, we randomly sample 10 instruction-answer pairs from all 4 test sets. Therefore, we evaluate 320 (8 x 4 x 10) datapoints in total for human evaluation. We do random sampling instead of evaluating all settings because hand evaluation of attributability is very costly and time-consuming (for examples of a source-sentence pair in the hand-evaluation, see \cref{lst:incorrectEntailment} or \cref{lst:partlyIncorrectEntailment}). In addition, the LLMs have uniform performance on different instructions. We also make all hand evaluations and LLM generations publicly available to justify our hand evaluation choice. Each sample is evaluated by one doctorate researcher. Given the extremely high overlaps in judgments in \cref{appendix:handEvalTraining} as well as the effort in manual annotation, we choose one annotator per sample to broaden the assessment spectrum.

GPT-4 evaluation is much less expensive than hand evaluation. We thus sample from all 14 raw models and fine-tuning settings. We also sample 25 instruction-answer pairs for each test set. Therefore, we evaluate (14 x 4 x 25) datapoints in total with GPT-4. \cref{tbl:humanGPTevalAtt} shows all settings in which we conduct human and GPT-4 evaluation. 

Finally, we aggregate all available scores to calculate Pearson correlations. For example, we aggregate 32 scores (8 settings x 4 test set) to compute the correlation between human and attributability scores. %The final verdicts are calculated in each setting and Pearson correlations are calculated to compare the different approaches.
As Table \ref{tab:validationAtt} shows, our \textit{answer attributability} score, the human and GPT-4 annotation arise at majorly the same results. This is signaled by correlation coefficients of over 80\%.

\begin{table}[t]
\centering
\scriptsize
\begin{tabular}{p{1.6cm}p{1cm}p{1cm}p{1cm}p{1cm}}
\hline
                                          & GPT-3.5-turbo   & GPT-4     & Llama-2-13b-chat & Zephyr-7b-$\beta$ \\ \hline
No Fine-Tuning                            & Hum./GPT & Hum./GPT & Hum./GPT        & Hum./GPT         \\
\textsc{SynSciQA}        & -         & -         & Hum./GPT              & Hum./GPT               \\
\textsc{SynSciQA}$_{S}$  & -         & -         & GPT              & GPT               \\
\textsc{SynSciQA+}       & -         & -         & GPT              & GPT               \\
\textsc{SynSciQA+}$_{S}$ & -         & -         & GPT              & GPT               \\
\textsc{SynSciQA++}      & -         & -         & Hum./GPT        & Hum./GPT         \\ \hline
Hum. = Human
\end{tabular}
\caption{Evaluation Settings for Annotating Instruction-Answers Pairs with Human (n=10 per setting) and GPT-4 (n=25 per setting) Annotation.}
\label{tbl:humanGPTevalAtt}
\end{table}

\section{Format Short-Cut in Attributability} \label{appendix:shortcut}
\begin{figure}[t]
	%\centering
	\includegraphics[width=\columnwidth]{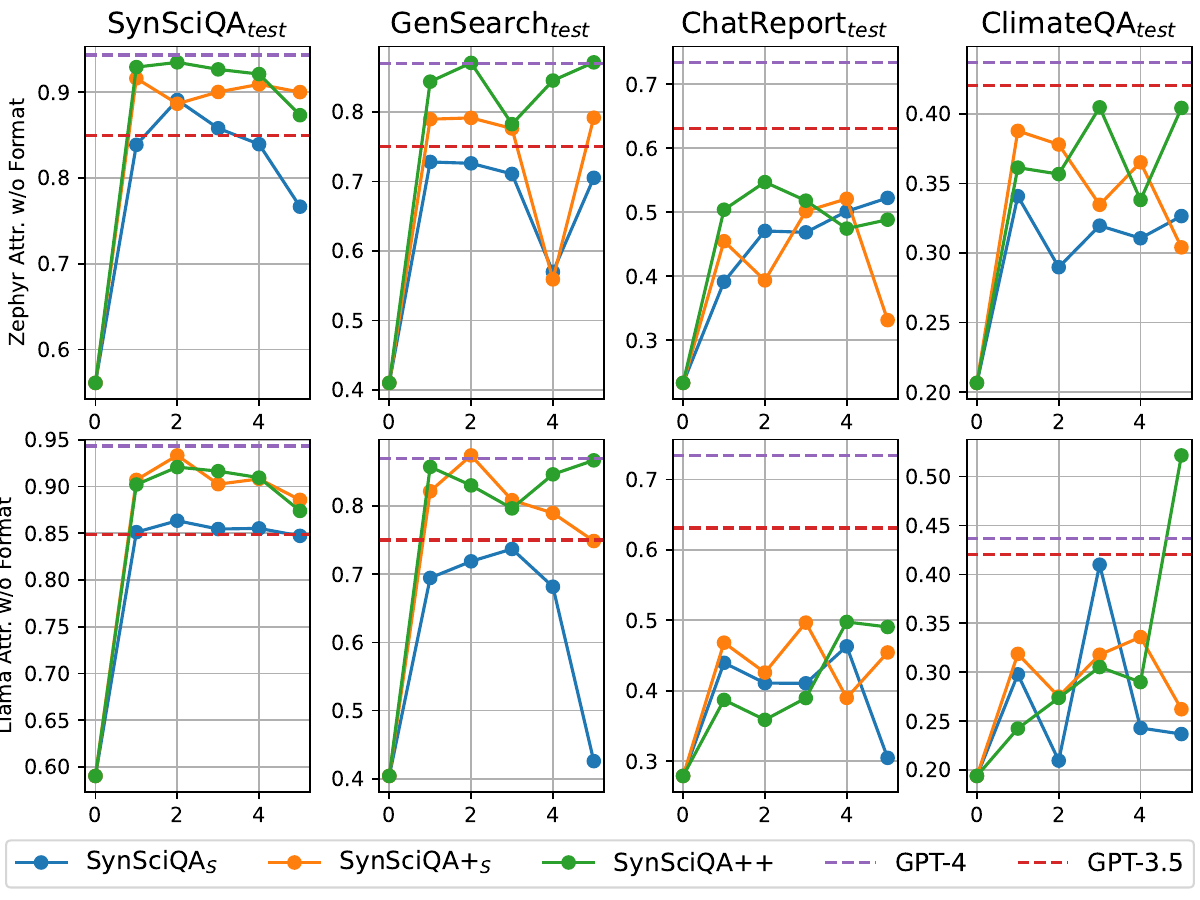}
  \captionsetup{skip=5pt}
	\caption{Controlling \textbf{quantity}, Attributability scores (without format-wrong sentences) vs. number of epoch, caused by different \textbf{quality}.}
	\label{fig:quality_attr_entail}
% \vspace{-0.7em}
\end{figure}

\begin{figure}[t]
	%\centering
	\includegraphics[width=\columnwidth]{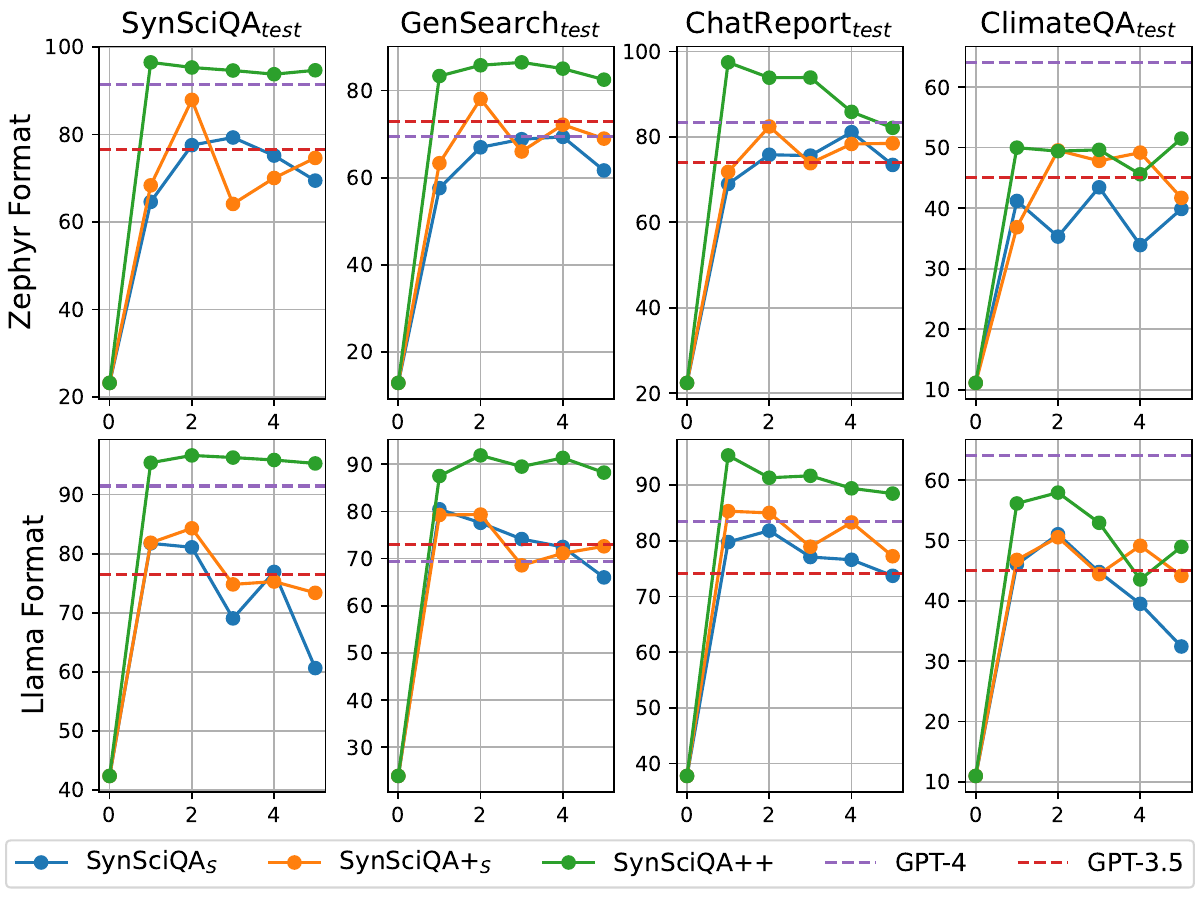}
  \captionsetup{skip=5pt}
	\caption{Controlling \textbf{quantity}, format correctness vs. number of epoch, caused by different \textbf{quality}.}
	\label{fig:quality_attr_format}
% \vspace{-0.7em}
\end{figure}

The fine-tuning may only improve format quality as a short-cut to improving attributability scores, instead of making the answer sentences more citation-compliant. To verify this, we compute the attributability score again on those format-correct sentences only. In other words, all improvements should then be caused by more sentences supported by sources. The results are shown in \cref{fig:quality_attr_entail}. It can be observed that in all settings, fine-tuning improves attributability without considering format quality. Interestingly, GPT-3.5 outperforms GPT-4 on \textsc{ChatReport}$_{test}$ ignoring format quality, which coheres to findings in \citet{ni-etal-2023-chatreport} where GPT-3.5 is better entailed in \textsc{ChatReport}$_{test}$. As a supplementary result, the improvement in format quality only is presented in \cref{fig:quality_attr_format}. Thus, the improvements in the Attributability score lie in both better formatting and entailment.

\end{document}